%% file: iclr2027_conference.tex
\documentclass{article} 
\usepackage{iclr2027_conference,times}
\iclrfinalcopy
\input{math_commands.tex}

\usepackage{bm}
\usepackage[T1]{fontenc}
\usepackage{hyperref}
\usepackage{url}

\usepackage{booktabs}
\usepackage{array}
\usepackage{multirow}
\usepackage[most]{tcolorbox}
\usepackage{appendix}
\usepackage{colortbl}
\usepackage{wrapfig}
\usepackage{algorithm}
\usepackage{algpseudocode}
\usepackage{titletoc}
\usepackage{fontawesome}
\usepackage{graphicx}
\usepackage{hyperref}
\usepackage{xcolor}
\usepackage{color}
\usepackage{enumitem}
\usepackage{cleveref}
\usepackage{arydshln}
\usepackage{tikz}
\crefformat{section}{\S#2#1#3}
\crefformat{subsection}{\S#2#1#3}
\crefformat{subsubsection}{\S#2#1#3}
\crefrangeformat{section}{\S\S#3#1#4 to~#5#2#6}
\crefmultiformat{section}{\S\S#2#1#3}{ and~#2#1#3}{, #2#1#3}{ and~#2#1#3}
\crefmultiformat{subsection}{\S\S#2#1#3}{ and~#2#1#3}{, #2#1#3}{ and~#2#1#3}
\Crefformat{figure}{#2Fig.~#1#3}
\Crefmultiformat{figure}{Figs.~#2#1#3}{ and~#2#1#3}{, #2#1#3}{ and~#2#1#3}
\Crefformat{table}{#2Tab.~#1#3}
\Crefmultiformat{table}{Tabs.~#2#1#3}{ and~#2#1#3}{, #2#1#3}{ and~#2#1#3}
\Crefformat{appendix}{Appx.~\S#2#1#3}
\crefmultiformat{appendix}{Appx.~\S#2#1#3}{ and~#2#1#3}{, #2#1#3}{ and~#2#1#3}
\crefformat{algorithm}{Alg.~#2#1#3}
\Crefformat{equation}{Eq.~#2#1#3}

\makeatletter
\newcommand*\myfontsize{%
  \@setfontsize\myfontsize{7pt}{8pt}%
}
\makeatother

\definecolor{myblue}{HTML}{054488}
\definecolor{mygreen}{HTML}{056B34}
\definecolor{myyellow}{HTML}{B8860B}
\definecolor{mygray}{HTML}{666666}
\definecolor{myred}{HTML}{B22222}

\newcommand{\mytextbox}[2]{%
  \begingroup
  \setlength{\fboxrule}{0.6pt}%
  \setlength{\fboxsep}{2pt}%
  \fcolorbox{#1}{white}{%
    {\myfontsize\bfseries\color{#1}#2}%
  }%
  \endgroup
}

\newcommand{\blue}[1]{\mytextbox{myblue}{#1}}
\newcommand{\green}[1]{\mytextbox{mygreen}{#1}}
\newcommand{\yellow}[1]{\mytextbox{myyellow}{#1}}
\newcommand{\gray}[1]{\mytextbox{mygray}{#1}}
\newcommand{\red}[1]{\mytextbox{myred}{#1}}

\hypersetup{
    colorlinks=true,
    linkcolor=blue,
    citecolor=blue,
}

\title{When Should Agents Think? Adaptive Reasoning via Cross-Turn Estimation}

\author{Yiruo Cheng$^1$\thanks{Work done during internship at Alibaba Token Hub, Alibaba Group} \ , Shen Huang$^2$, Xiaoshuai Song$^1$, Jiejun Tan$^1$, Guanting Dong$^1$, \\
\textbf{Pengjun Xie$^2$, Ji-Rong Wen$^1$, Zhicheng Dou$^1$\thanks{Corresponding author.}} \\
$^1$Gaoling School of Artificial Intelligence,
Renmin University of China
\\
$^2$Alibaba Token Hub, Alibaba Group \\
\texttt{\{chengyr, dou\}@ruc.edu.cn}\\
\\
\begin{tabular}{@{}ll@{}}
\faGithub\ GitHub: \href{https://github.com/HututuAI/RACE}{\texttt{\textcolor{cyan}{https://github.com/HututuAI/RACE}}}
\end{tabular}
}

\begin{document}

\maketitle

\begin{abstract}
Large language model (LLM)-based agents have demonstrated strong capabilities on complex tasks. They typically perform reasoning before each action throughout an interaction trajectory. However, reasoning may not be necessary at every turn, as reasoning produced earlier can continue to support subsequent actions. A key challenge is therefore to determine when existing reasoning remains sufficient and when a new reasoning step is needed, without relying on costly generation-based verification. We find that decreases in the likelihood of subsequent reference actions after removing additional reasoning closely track whether those actions remain recoverable given earlier reasoning, providing an effective and lightweight signal for estimating cross-turn action support. Based on this observation, we propose \textbf{R}easoning \textbf{A}daptation through \textbf{C}ross-Turn \textbf{E}stimation (\textbf{RACE}), a training approach for adaptive agent reasoning. RACE introduces a \textbf{L}ikelih\textbf{o}od-\textbf{G}uided Progressive Reason\textbf{i}ng \textbf{C}over Detection \textbf{(LoGiC)} procedure that progressively identifies reasoning turns whose removal has limited impact on the current and subsequent reference actions. The resulting removal signals are incorporated into both supervised fine-tuning and agentic reinforcement learning, enabling the policy to learn when to reason and when to act directly. Extensive experiments on four representative agent benchmarks show that RACE substantially reduces reasoning cost while maintaining or improving task performance. 

\end{abstract}

\section{INTRODUCTION}
Recent advances in Large Reasoning Models (LRMs)~\citep{openai_o1,deepseek_r1,qwq32b} have significantly improved the capabilities of agents on complex tasks that require sustained planning and execution across multiple interaction turns~\citep{ragen,openresearcher,htr,deepagent,hiersearch}.
In these tasks, agents make sequential decisions by incorporating historical actions and environment feedback into their evolving context. 
Under this paradigm, many existing agents explicitly generate reasoning steps before action decisions~\citep{react,webthinker,chatshopbuddy,deepresearcher}, 
implicitly treating reasoning as necessary at every interaction turn and incurring substantial inference cost over long trajectories~\citep{tool_light,ares,search_reason}.

\begin{figure*}[t]
    \centering 
    \includegraphics[width=0.98\linewidth]
    {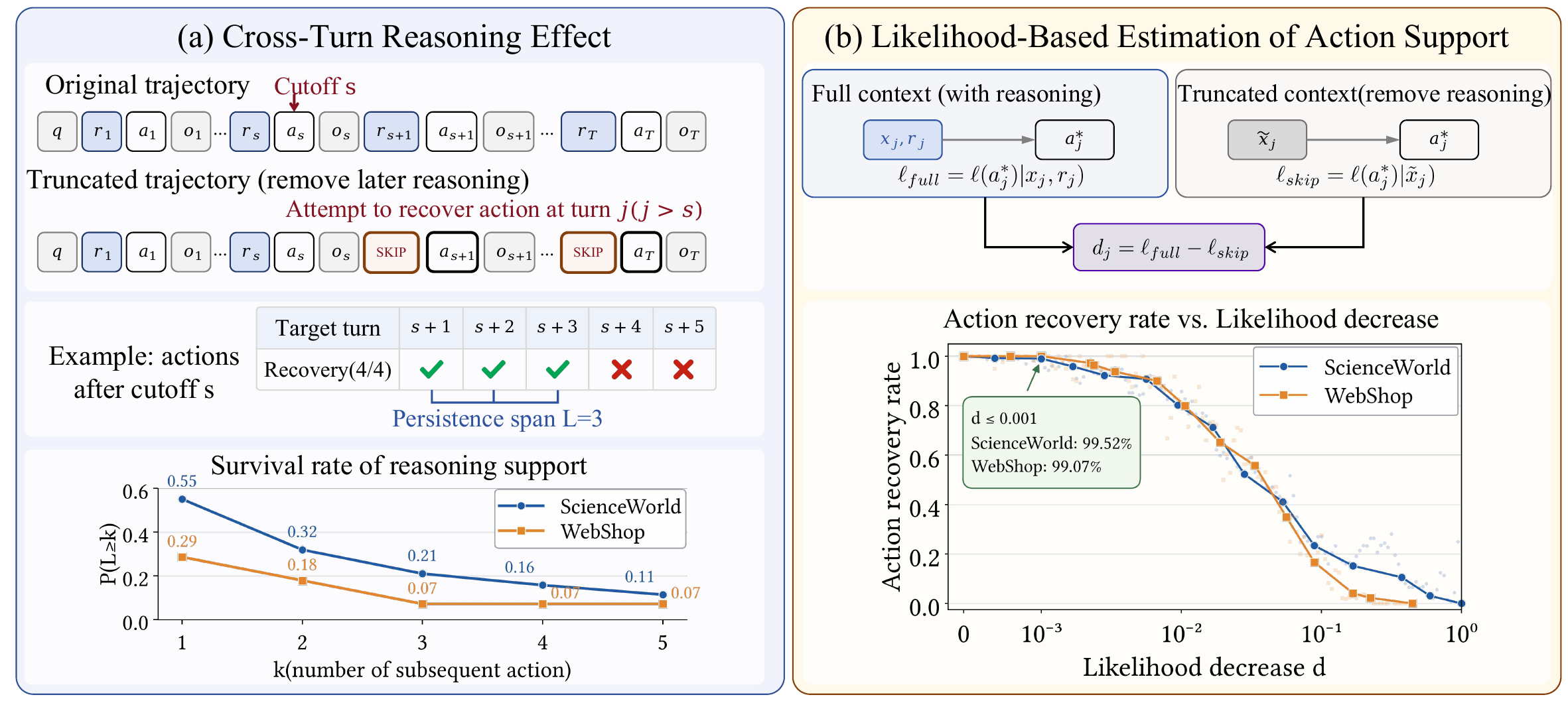}
    \caption{Analysis of cross-turn reasoning support. (a) Reference-action recovery at subsequent turns after removing later reasoning. (b) Decreases in the likelihood of subsequent reference actions closely correspond to the recovery rate, providing a lightweight proxy for action recoverability.}
    \label{fig:preliminary}
\end{figure*}

However, reasoning may not be confined to the interaction turn in which it is generated. Our analysis shows that earlier reasoning can support subsequent actions over varying ranges: some reasoning steps remain useful for only one subsequent turn, while others continue to support actions across multiple turns, as shown in Figure~\ref{fig:preliminary}. 
Such persistence may be explained by the fact that previously generated plans, constraints, or intermediate judgments can remain useful beyond their original turns~\citep{agent_omit} and continue to support subsequent actions. 
We refer to this as cross-turn reasoning persistence. 
Under this perspective, we think agents do not need to generate reasoning at every turn when previous reasoning remains sufficient to support later actions. A new reasoning step is only required when changes in the interaction context make previous reasoning insufficient. 

This perspective aligns with recent studies on \emph{adaptive reasoning}~\citep{adaptthink,ares,dart,agent_omit}, which seek to reduce inference cost by determining whether reasoning is necessary at each interaction turn.
To obtain turn-level removal signals, they typically compare actions generated with and without a reasoning step~\citep{CogRouter}.
However, these approaches do not explicitly account for the varying cross-turn persistence of reasoning, with reasoning necessity typically assessed independently at each turn.
Moreover, obtaining the removal signals requires repeated generations for candidate reasoning turns, increasing the cost of supervised fine-tuning data synthesis as trajectories grow longer~\citep{steer}. 
This challenge becomes more pronounced in on-policy reinforcement learning, where generation-based verification must be repeatedly performed at every interaction turn to obtain local turn-level signals for newly sampled trajectories as the policy evolves, introducing substantial computational overhead and limiting scalability.
Therefore, the key to reducing inference cost is not independently determining whether reasoning steps can be removed, but estimating how far existing reasoning remains sufficient and when a new reasoning step becomes necessary.

In this paper, we investigate this problem and seek a lightweight alternative to turn-level verification of reasoning necessity. We find that decreases in the likelihood of subsequent reference actions after removing additional reasoning closely track whether those actions remain recoverable given earlier reasoning. This observation motivates using likelihood decrease of subsequent actions as a lightweight signal for evaluating reasoning removal from existing trajectories, without repeatedly generating candidate actions for verification.
Therefore, we propose \textbf{RACE} (\textbf{R}easoning \textbf{A}daptation through \textbf{C}ross-Turn \textbf{E}stimation), a training approach that learns adaptive reasoning behavior by estimating the cross-turn impact of reasoning removal.
The core of RACE is a \textbf{L}ikelih\textbf{o}od-\textbf{G}uided Progressive Reason\textbf{i}ng \textbf{C}over Detection \textbf{(LoGiC)} procedure that uses reference-action likelihood decrease as a lightweight signal for reasoning removal.
In detail, LoGiC assesses each candidate reasoning by measuring the effect of its removal on the likelihood of the current and subsequent reference actions under the resulting context. A candidate is removed only when the decrease remains within a predefined threshold across all affected actions. The retained reasoning steps collectively form a \emph{reasoning cover} that provides sufficient support for actions throughout the trajectory.
Unlike generation-based turn-level verification, LoGiC requires only likelihood computation for existing reference actions, enabling efficient derivation of turn-level reasoning removal signals for training.

RACE further incorporates these signals into both supervised fine-tuning (SFT) and agentic reinforcement learning (RL). During the SFT stage, LoGiC transforms full-reasoning trajectories into adaptive reasoning trajectories, teaching the model when it can act directly without additional reasoning. Building on this initialization, RACE introduces cover-aware agentic RL, where LoGiC is applied to successful trajectories sampled from the current policy to derive their reasoning removal signals. The signals are converted into reasoning-closure supervision, providing a direct learning signal for when to terminate reasoning. In this way, the task reward optimizes task performance, while LoGiC-derived reasoning removal signals shape adaptive reasoning behavior.

We evaluate RACE on four representative agent benchmarks spanning scientific reasoning, web shopping, application use, and knowledge-intensive reasoning. 
Compared with standard SFT followed by RL, RACE maintains comparable task performance while reducing reasoning tokens by 32.3\%, 80.2\%, 46.4\%, and 62.0\% on ScienceWorld, WebShop, AppWorld, and DeepSearch, validating its ability to learn effective adaptive reasoning behavior.

Our main contributions are summarized as follows:

\noindent $\bullet$ We characterize the cross-turn effect of agent reasoning, showing that earlier reasoning can continue to support subsequent actions over varying ranges, and that reference-action likelihood decrease provides a lightweight proxy for estimating this support.

\noindent $\bullet$  We propose \textbf{RACE}, a training approach for adaptive agent reasoning that leverages likelihood-guided cross-turn estimation to derive reasoning removal signals and incorporates them into both supervised fine-tuning and agentic reinforcement learning.

\noindent $\bullet$  Extensive experiments on four representative agent benchmarks show that RACE substantially reduces reasoning cost while maintaining or improving task performance.

\section{PRELIMINARIES}
\label{sec:preliminaries}
Before presenting our method, we first establish the problem formulation, measure how far existing reasoning can support subsequent actions through repeated action generation, and further seek a lightweight signal that does not require repeated action generation. 
\subsection{PROBLEM FORMULATION}
\label{sec:problem_formulation}
To formalize our study of agent reasoning, we consider an agent completing a task $q$ over $T$ decision turns. 
At each turn $t$, it takes action $a_t$ and receives observation $o_t$, with optional reasoning $r_t$ generated before the action.
Let $x_t=(q,r_1,a_1,o_1,\ldots,a_{t-1},o_{t-1})$ denote the context before turn $t$. 
Rather than requiring reasoning before every action, 
we consider two execution modes under the policy $\pi_\theta$. At turn $t$, the agent may generate reasoning and then take an action,
\begin{align}
r_t\sim\pi_\theta(\cdot\mid x_t),
\qquad
a_t\sim\pi_\theta(\cdot\mid x_t,r_t),
\end{align}
or act directly without generating additional reasoning at the current turn,
\begin{align}
a_t \sim \pi_\theta(\cdot \mid x_t).
\end{align}

Our objective is to optimize $\pi_\theta$ to selectively invoke reasoning across decision turns while still preserving task performance during agent execution.

\subsection{Measuring Cross-Turn Support through Generation}
\label{sec:reasoning_persistence}
To directly measure how far existing reasoning can support subsequent actions, we analyze successful trajectories generated by Qwen3.5-9B on ScienceWorld and WebShop.
Given a successful full-reasoning trajectory
$\tau=(q,(r_t,a_t^*,o_t)_{t=1}^{T})$, where $a_t^*$ is the reference action at turn $t$, we select turn $s$ as the cutoff, retain all reasoning through turn $s$, and remove all \textbf{reasoning} steps after turn $s$ while keeping the original actions and observations unchanged.

For each turn $j>s$, we denote the resulting truncated context as $\tilde{x}_j$, generate $K$ actions from this context, and compute the reference-action recovery rate:
\begin{align}
\operatorname{Rec}_j
=
\frac{1}{K}
\sum_{k=1}^{K}
\mathbb{I}
\left[
\hat{a}_j^{(k)} \simeq a_j^*
\right],
\qquad
\hat{a}_j^{(k)}
\sim
\pi_\theta(\cdot \mid \tilde{x}_j).
\end{align}
Here, $K=4$, and $\simeq$ denotes canonical action equivalence.
As shown in Figure~\ref{fig:preliminary}(a), reasoning from earlier turns can continue to support later actions, allowing subsequent actions to be recovered without additional reasoning. However, the effective range of this support varies: some reasoning supports only the immediate action, while others remain useful across multiple subsequent turns.

While reference-action recovery rate provides a direct measure of such support, it requires full autoregressive decoding for every candidate context. This makes repeated verification costly for large-scale data synthesis and especially for on-policy reinforcement learning.

\subsection{LIKELIHOOD-BASED ESTIMATION OF ACTION SUPPORT}
\label{sec:action_support}
Generation-based recovery rate directly measures whether subsequent reference actions remain recoverable when only earlier reasoning is retained, while this behavior may also be reflected in their likelihoods. If a subsequent action remains recoverable given earlier reasoning, its reference-action likelihood should remain relatively stable; otherwise, it should decrease. We therefore examine the relationship between likelihood decrease and generation-based recovery rate.

For a reference action $a_j^*$ with $M_j$ tokens, we define its mean token log-likelihood under context $h$:
\begin{align}
\ell(a_j^*\mid h)
=
\frac{1}{M_j}
\sum_{m=1}^{M_j}
\log \pi_\theta
\left(
a_{j,m}^*
\mid
h,a_{j,<m}^*
\right).
\label{eq:mean_action_log}
\end{align}
For each truncated context $\tilde{x}_j$ and the corresponding complete context $(x_j,r_j)$, we define the decrease in the reference action likelihood as
\begin{align}
d_j
=
\ell(a_j^*\mid (x_j,r_j))
-
\ell(a_j^* \mid \tilde{x}_j).
\end{align}

As shown in Figure~\ref{fig:preliminary}(b), reference-action likelihood decrease closely corresponds to reference-action recovery rate. Small decreases generally correspond to reliable recovery, while larger decreases are associated with lower recovery rates. This suggests that likelihood decrease can serve as a lightweight proxy for subsequent action recoverability given earlier reasoning. Unlike action generation, it only requires batched teacher-forced scoring of reference actions.

\section{Methodology}
\begin{figure*}[t]
    \centering 
    \includegraphics[width=0.98\linewidth]
    {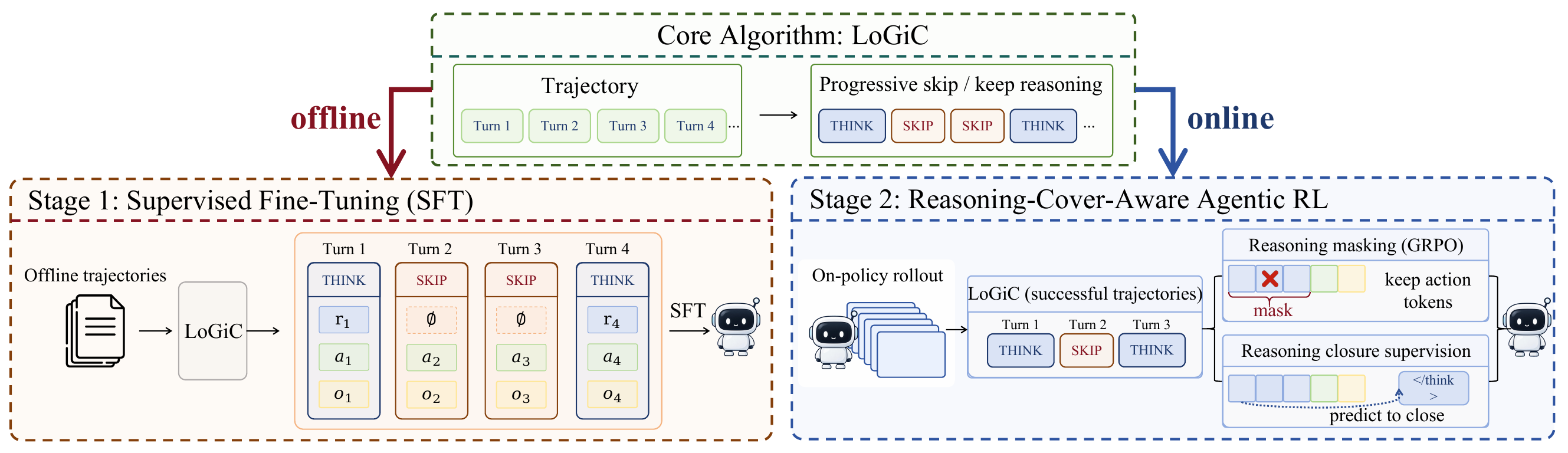}
    \caption{Overview of our training approach RACE. RACE uses the LoGiC algorithm to identify reasoning steps that can be safely removed, and the resulting signals are used in both supervised fine-tuning and agentic reinforcement learning to learn adaptive reasoning behavior.}
    \label{fig:RACE}
\end{figure*}

In this section, we introduce RACE, a training approach that enables agents to adaptively determine when to reason and when to act directly. An overview of RACE is illustrated in Figure~\ref{fig:RACE}.

\begin{itemize}[leftmargin=1em]
\item \textbf{Likelihood-Guided Progressive Reasoning Cover Detection} (\cref{sec:LoGiC}):
We use reference-action likelihood decrease to identify removable reasoning and construct a reasoning cover.

\item \textbf{Agent Reasoning Synthesis for Supervised Fine-Tuning} (\cref{sec:behavior-synthesis}): Based on the identified reasoning cover, we synthesize adaptive reasoning trajectories for supervised fine-tuning.

\item \textbf{Reasoning-Cover-Aware Agentic Reinforcement Learning} (\cref{sec:cover-aware-rl}):
We further develop a cover-aware agentic RL approach that masks removable reasoning from policy-gradient optimization and provides reasoning-closure supervision on successful rollout trajectories.

\end{itemize}

\begin{figure*}[t]
    \centering 
    \includegraphics[width=0.98\linewidth]
    {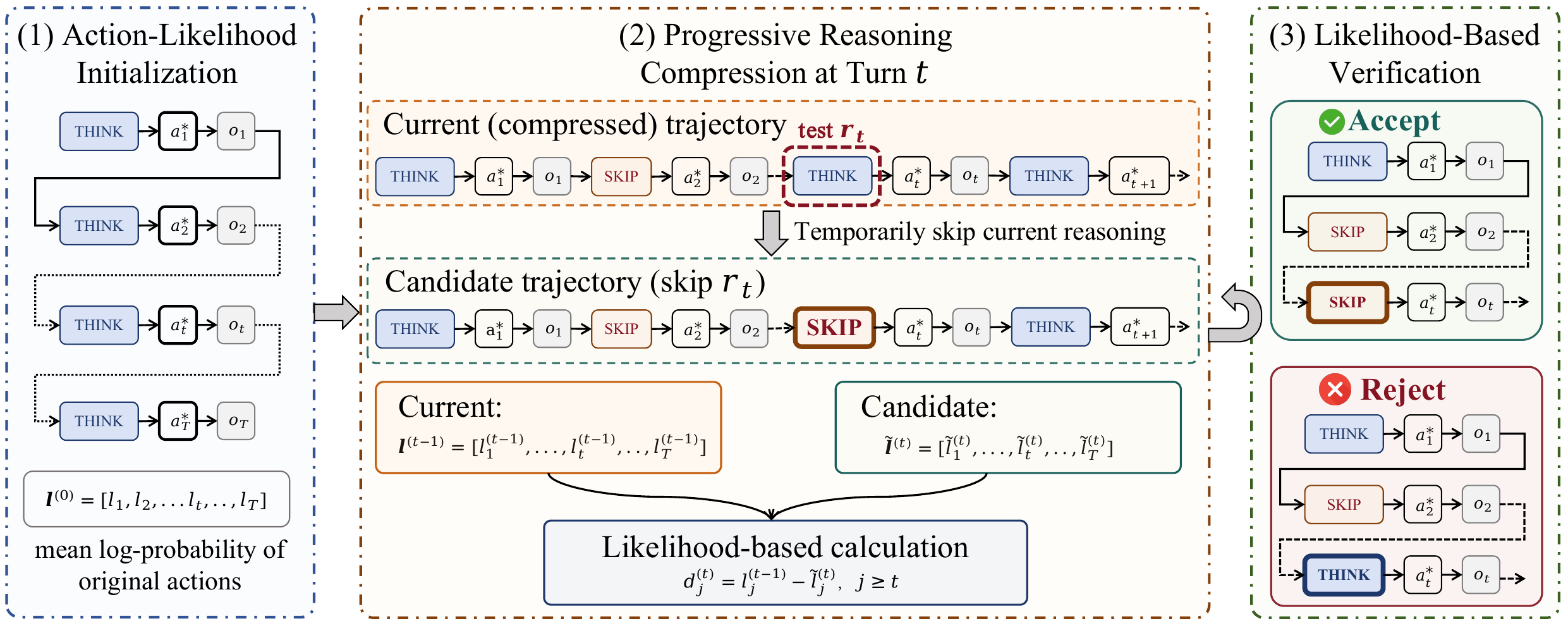}
    \caption{Overview of the Likelihood-Guided Progressive Reasoning Cover Detection (LoGiC). (1) Reference-action likelihoods are initialized on the full trajectory. Starting from the second turn, (2) LoGiC tests whether the reasoning can be removed, and (3) verifies the removal based on likelihood decrease, with each accepted removal updating the context used to evaluate subsequent turns.}
    \label{fig:LoGiC}
\end{figure*}

\subsection{Likelihood-Guided Progressive Reasoning Cover Detection}
\label{sec:LoGiC}
Motivated by the correspondence between likelihood decrease and action recovery observed in~\cref{sec:action_support}, we use reference-action likelihood decrease to evaluate the effect of reasoning removals and progressively construct a reasoning cover.
As shown in Figure~\ref{fig:LoGiC}, the procedure consists of three steps: 

\textbf{(1) Action-Likelihood Initialization.}
Given an agent trajectory $\tau=(q,(r_t,a_t,o_t)_{t=1}^{T})$, we treat each action $a_t$ as a fixed reference action, denoted by $a_t^*$, for likelihood evaluation under its original full-reasoning context.
Following Equation~\ref{eq:mean_action_log}, we initialize the action-likelihood vector as
\begin{align}
\bm{\ell}^{(0)}
=
\left[
\ell(a_1^* \mid x_1,r_1),
\ldots,
\ell(a_T^* \mid x_T,r_T)
\right].
\end{align}
This vector serves as the reference for subsequent reasoning-cover construction.

\textbf{(2) Progressive Reasoning Compression.}
To assess the effect of removing a reasoning step under the current compressed context, we construct a candidate trajectory without that reasoning and recompute the likelihoods of the current and subsequent reference actions for comparison.

Specifically, we evaluate each reasoning step sequentially starting from the second turn and carry accepted removals forward to subsequent turns. 
Let $x^{(t)}$ denote the context before turn $t$ after applying all reasoning removals accepted at preceding turns. In particular,
$x^{(1)}=(q)$ and
$x^{(2)}=\left(x^{(1)},r_1,a_1^*,o_1\right)$.
At turn $t\in\{2,\ldots,T\}$, we remove $r_t$ to construct a candidate trajectory and compute the corresponding candidate action-likelihood vector as
\begin{equation}
\widetilde{\bm{\ell}}^{(t)}
=
\left[
\widetilde{\ell}_1^{(t)},
\ldots,
\widetilde{\ell}_T^{(t)}
\right],
\end{equation}
where
\begin{align}
\widetilde{\ell}_j^{(t)}
=
\begin{cases}
\ell_j^{(t-1)},
& j<t, \\[2mm]
\ell(a_t^* \mid x^{(t)}),
& j=t, \\[2mm]
\ell(a_j^* \mid x^{(t)}, a_t^*, o_t,
(r_u,a_u^*,o_u)_{u=t+1}^{j-1},
r_j),
& j>t.
\end{cases}
\end{align}
In the case of $j>t$ above, the candidate trajectory differs from the current trajectory only by the removal of $r_t$, while all subsequent reasoning, actions, and observations remain unchanged.

\textbf{(3) Likelihood-Based Verification.}
To determine whether a candidate reasoning step can be safely removed under the current compressed context, we measure its effect on the likelihoods of the current and subsequent reference actions.
Specifically, for each reference action $a_j^*$ for $j\in\{t,\ldots,T\}$, we compute the corresponding likelihood decrease as
\begin{align}
d_j^{(t)}
=
\ell_j^{(t-1)}
-
\widetilde{\ell}_j^{(t)}.
\end{align}
We accept the removal of $r_t$ only if the decrease remains within the threshold $\epsilon$ consistently for all current and subsequent reference actions throughout the entire trajectory:
\begin{align}
d_j^{(t)}
\le
\epsilon,
\qquad
\forall j\in\{t,\ldots,T\}.
\end{align}
The likelihood vector and context are then updated according to the verification result:
\begin{align}
\left(
\bm{\ell}^{(t)},
x^{(t+1)}
\right)
=
\begin{cases}
\left(
\widetilde{\bm{\ell}}^{(t)},
\left(x^{(t)},a_t^*,o_t\right)
\right),
& \text{if accepted},\\[2mm]
\left(
\bm{\ell}^{(t-1)},
\left(x^{(t)},r_t,a_t^*,o_t\right)
\right),
& \text{otherwise}.
\end{cases}
\end{align}
After all turns have been examined, the retained reasoning steps constitute the reasoning cover of $\tau$.

\subsection{Agent Reasoning Synthesis for Supervised Fine-Tuning}
\label{sec:behavior-synthesis}

To enable agents to adaptively determine when to reason and when to act directly, we use the reasoning covers identified by the Likelihood-Guided Progressive Reasoning Cover Detection (LoGiC) algorithm in \cref{sec:LoGiC} to synthesize adaptive reasoning trajectories for supervised fine-tuning.

\paragraph{LoGiC-Based Data Synthesis.}
Given an agent trajectory $\tau$, we apply the LoGiC algorithm to identify reasoning that can be removed and obtain the corresponding adaptive agent reasoning trajectory
$
\tilde{\tau}=\operatorname{LoGiC}(\tau)
$.
The resulting trajectories constitute the SFT training data $\mathcal{D}_{\mathrm{SFT}}$.

\paragraph{Supervised Fine-Tuning.}
We fine-tune the agent on $\mathcal{D}_{\mathrm{SFT}}$ using supervised learning:
\begin{align}
\mathcal{L}_{\mathrm{SFT}}
=
-
\mathbb{E}_{\tilde{\tau}\sim\mathcal{D}_{\mathrm{SFT}}}
\left[
\sum_{n=1}^{|\tilde{\tau}|}
\log
\pi_\theta
\left(
\tilde{\tau}_n
\mid
\tilde{\tau}_{<n}
\right)
\right].
\end{align}
The loss is applied only to model-generated reasoning and action tokens. This stage provides direct supervision for learning when to generate reasoning and when to act directly, serving as an initialization for subsequent agentic reinforcement learning.

\subsection{REASONING-COVER-AWARE AGENTIC REINFORCEMENT LEARNING}
\label{sec:cover-aware-rl}

Supervised fine-tuning provides an initial adaptive reasoning policy from fixed offline trajectories, which may not match the evolving policy. We therefore incorporate LoGiC into on-policy agentic reinforcement learning. For each rollout batch, LoGiC identifies removable reasoning steps in successful trajectories to mask unnecessary reasoning from the policy-gradient objective and supervise reasoning termination. Failed trajectories follow standard GRPO.

\paragraph{LoGiC-Based Reasoning Masking.}
We first prevent GRPO from reinforcing reasoning that LoGiC identifies as unnecessary. Specifically, for each successful on-policy trajectory, LoGiC derives a set of SKIP turns $S_i$. We mask the reasoning tokens at these turns from the GRPO objective while retaining their action tokens for task optimization. Let $\mathcal{P}_i$ denote all policy-token positions in $\tau_i$, and $\mathcal{K}_i\subseteq\mathcal{P}_i$ denote those remaining after masking turns in $S_i$. The masked GRPO loss is
\begin{equation}
\mathcal{L}_{\mathrm{GRPO}}
=
-\frac{1}{
\sum_{i\in\mathcal{B}} |\mathcal{P}_i|
}
\sum_{i\in\mathcal{B}}
\sum_{k\in\mathcal{K}_i}
\min
\left(
r_{i,k}(\theta)\widehat{A}_i,\,
\operatorname{clip}
\left(
r_{i,k}(\theta),
1-\epsilon_{\mathrm{clip}},
1+\epsilon_{\mathrm{clip}}
\right)
\widehat{A}_i
\right).
\end{equation}
Here, $r_{i,k}(\theta)$ is the importance ratio and $\widehat{A}_i$ the group-relative advantage. Masked reasoning tokens are excluded from the policy gradient while their original positions remain in the normalization.

\paragraph{LoGiC-Based Reasoning Closure.}
Masking prevents redundant reasoning from being reinforced, but does not directly encourage the policy to terminate reasoning at these turns. 
We therefore introduce reasoning-closure supervision on LoGiC-verified SKIP turns. 
Specifically, let $e$ denote the reasoning termination token, $\mathcal{I}_{\mathrm{succ}}$ denote the successful trajectories, $S_i$ denote their LoGiC-verified SKIP turns, and $x_i^{(t)}$ denote the corresponding compressed context. The reasoning-closure loss is
\begin{equation}
\mathcal{L}_{\mathrm{close}}
=
-\frac{1}{\sum_{i\in\mathcal{B}} T_i}
\sum_{i\in\mathcal{I}_{\mathrm{succ}}}
\sum_{t\in S_i}
\log
\pi_\theta
\left(
e\mid x_i^{(t)}
\right),
\end{equation}
where $T_i$ is the number of interaction turns in $\tau_i$. Normalizing by total rollout turns makes closure supervision scale with the prevalence of LoGiC-verified SKIP turns.
The final actor objective combines task optimization with LoGiC-Based reasoning closure:
\begin{equation}
\mathcal{L}_{\mathrm{actor}}
=
\mathcal{L}_{\mathrm{GRPO}}
+
\lambda \mathcal{L}_{\mathrm{close}},
\end{equation}
where $\lambda$ controls the strength of reasoning-closure supervision.

\section{Experiments}
\subsection{Experimental Settings}

\begin{table*}[!t]
\centering
\caption{Main results on four agent tasks, including ScienceWorld, WebShop, AppWorld, and DeepSearch.
\textit{Avg Tok.} denotes the average number of reasoning tokens generated per trajectory.
For models with up to 32B parameters, the best results are in \textbf{bold} and the second are \underline{underlined}. 
}
\label{tab:reasoning_performance}
\setlength\tabcolsep{2pt}
\fontsize{8pt}{11.5pt}\selectfont
\begin{tabular}{p{3.2cm}*{8}{c}}
\toprule
\multirow{2}[2]{*}{\textbf{Method}} &  \multicolumn{2}{c}{\textbf{ScienceWorld}} & \multicolumn{2}{c}{\textbf{WebShop}} & \multicolumn{2}{c}{\textbf{AppWorld}} & \multicolumn{2}{c}{\textbf{DeepSearch}} \\
\cmidrule(lr){2-3} \cmidrule(lr){4-5} \cmidrule(lr){6-7}  \cmidrule(lr){8-9}
 &   Success $\uparrow$ & Avg Tok.$\downarrow$  & Success $\uparrow$ & Avg Tok.$\downarrow$  & Success $\uparrow$ & Avg Tok. $\downarrow$ & Success $\uparrow$ & Avg Tok.$\downarrow$ \\
\midrule
\multicolumn{9}{l}{\textit{\textbf{ReAct-style Iterative Reasoning Agents}}} \\

DeepSeek-V4-Pro & 32.11 & 1,662 & 40.20 & 1,288 & 87.69 & 3,016 & 40.00 & 1,243\\
GLM-5.3 & 30.29 & 1,854 & 46.60 & 1,715 & 77.61  & 17,142   & 24.00 & 3,287 \\
GPT-5.5 (Medium) & 69.88 & 793 & 40.20 & 799 & 91.11 & 1,002 &46.25 & 557 \\
Claude-Opus-4.6 (Medium) & 59.37 & - & 53.40 & - &78.80 & -&30.00&-\\
Kimi-K3 & 45.52 & 1,633 & 45.80 & 1,471 & 80.00 & 2,751 &44.25 & 1,679\\
Qwen3.8-27B & 38.37& 2,011 & 35.60 & 1,059&58.46&4,425 & 33.50 & 2,623 \\
Qwen3.5-9B & 18.52 & 2,210 & 15.60 & 2,713 &46.15&3,134 & 29.25 & 1,281\\

\midrule
\multicolumn{9}{l}{\textit{\textbf{Reasoning-Efficient Agents (Qwen3.5-9B)}}} \\
Think Tool &19.63 &2,426&14.00&2,154& 43.59 & 2,943 & 25.75 & \underline{987}\\
ADaPT & 14.24 & 5,512 & 31.80 & 3,720 & 21.20 &4,166 & 25.75 & 1,797\\
Sparse ReAct & 19.85 & 2,126 & 16.40 & 2,426 & 45.47& 3,187 & 26.00 & 2,228\\
DART & 17.92 & 1,413& 18.20&1,791& 49.06  &\underline{2,345} & 29.00 & 1,316\\
AdaptThink & \underline{64.49} & \textbf{1,252} & 44.60 & \underline{748} &\textbf{68.72}& 2,381&\underline{33.50}&1,481\\
\rowcolor[RGB]{236,244,252}
RACE-SFT & 43.05 & 1,966 & \underline{45.40} & 1,232 & 63.08 & 3,330 & 28.25 & 1,856\\
\rowcolor[RGB]{236,244,252}
RACE-SFT-RL & \textbf{73.28} & \underline{1,403} & \textbf{53.40} & \textbf{535} & \underline{65.13} & \textbf{1,900} & \textbf{35.00} & \textbf{822}\\

\bottomrule
\end{tabular}
\end{table*}

\textbf{Datasets and Evaluation.}
We evaluate RACE on four representative agent benchmarks:
(1) \textbf{Scientific Reasoning}: ScienceWorld~\citep{sciworld} evaluates science tasks in simulated environments requiring sequential reasoning and actions.
(2) \textbf{Web Shopping}: WebShop~\citep{webshop} evaluates product search and selection based on user requirements.
(3) \textbf{Application Use}: AppWorld~\citep{appworld} evaluates tasks through interactions with simulated apps and APIs.
(4) \textbf{Knowledge-Intensive Reasoning}: DeepSearch~\citep{search-r1} evaluates knowledge-intensive question answering through retrieval and reasoning.
We report success rate and average reasoning tokens per trajectory. Dataset and implementation details are in Appendices~\ref{sec:datasets_details} and~\ref{sec:implementation_details}.

\textbf{Baselines.}
We compare RACE with the following agent reasoning baselines:
\textbf{(1) ReAct-style Iterative Reasoning Agents.}
We evaluate ReAct~\citep{react}-style reasoning agents across a range of representative open-source and closed-source models, including DeepSeek-V4-Pro~\citep{deepseek-v4}, GLM-5.3~\citep{glm-5}, GPT-5.5~\citep{openai2026gpt55}, Claude-Opus-4.6~\citep{anthropic2026claudeopus46}, Kimi-K3~\citep{kimik3}, and Qwen3.8 series~\citep{qwen3.8}.
\textbf{(2) Reasoning-Efficient Agents.}
We further compare with Think Tool~\citep{anthropic_think_tool},
ADaPT~\citep{adapt}, Sparse ReAct~\citep{react},  DART~\citep{dart}, and AdaptThink~\citep{adaptthink}. 
All reasoning-efficiency methods are evaluated with Qwen3.5-9B~\citep{qwen3.5} as the backbone to ensure a fair comparison. 
Detailed descriptions of these baselines are provided in Appendix~\ref{sec:baseline_details}.

\subsection{main results}
Table~\ref{tab:reasoning_performance} presents the main results across four representative agent benchmarks, evaluating both task performance and reasoning cost. We summarize the key findings below.

\textbf{Competitive Performance against Strong LLM Agents.}
RACE achieves competitive performance against substantially larger and proprietary LLM agents, with success rates of 73.28, 53.40, 65.13, and 35.00 on ScienceWorld, WebShop, AppWorld, and DeepSearch, respectively. Interestingly, several LLM agents sometimes take actions without additional explicit reasoning, suggesting that such adaptive behavior can emerge in capable agents. RACE learns this behavior through training, skipping reasoning when appropriate while maintaining performance across diverse agent tasks.

\textbf{Superior Performance and Reasoning Efficiency.}
RACE achieves a strong performance-efficiency trade-off across four agent benchmarks.
Compared with Qwen3.5-9B, RACE-SFT-RL improves success by 54.76, 37.80, 18.98, and 5.75 points while reducing reasoning tokens by 36.5\%, 80.3\%, 39.4\%, and 35.8\% on ScienceWorld, WebShop, AppWorld, and DeepSearch, respectively.
Among same-backbone baselines, RACE-SFT-RL performs best on ScienceWorld, WebShop, and DeepSearch, while using fewer tokens than AdaptThink on WebShop, AppWorld, and DeepSearch.

\textbf{Further Gains from Cover-Aware Reinforcement Learning.}
Cover-aware RL further improves upon RACE-SFT. Compared with RACE-SFT, RACE-SFT-RL increases success by 30.23, 8.00, 2.05, and 6.75 points on ScienceWorld, WebShop, AppWorld, and DeepSearch, respectively, while reducing reasoning tokens by 28.6\%, 56.6\%, 42.9\%, and 55.7\%. These simultaneous gains show that LoGiC serves as an effective proxy for deriving turn-level reasoning removal signals during policy optimization without repeated generation-based verification.

\begin{table*}[t]
\centering
\caption{Ablation studies of RACE. The best results are in \textbf{bold}. 
}
\label{tab:abalation}
\setlength\tabcolsep{0.2pt}
\fontsize{8pt}{11pt}\selectfont
\begin{tabular}{p{4.6cm}*{8}{c}}
\toprule
\multirow{2}[2]{*}{\textbf{Method}} &  \multicolumn{2}{c}{\textbf{ScienceWorld}} & \multicolumn{2}{c}{\textbf{WebShop}} & \multicolumn{2}{c}{\textbf{AppWorld}} & \multicolumn{2}{c}{\textbf{DeepSearch}} \\
\cmidrule(lr){2-3} \cmidrule(lr){4-5} \cmidrule(lr){6-7}  \cmidrule(lr){8-9}
 &   Success $\uparrow$  & Avg Tok. $\downarrow$  & Success $\uparrow$ & Avg Tok. $\downarrow$  & Success $\uparrow$  & Avg Tok.$\downarrow$  & Success $\uparrow$ & Avg Tok.$\downarrow$ \\
\midrule

\rowcolor[RGB]{236,244,252}
RACE-SFT-RL & 73.28 & 1,403 & \textbf{53.40} & 535 &  65.13 & \textbf{1,900} & 35.00 & \textbf{822}\\

\hspace{0.1cm}w/o Reasoning Masking in RL & 72.95 & \textbf{609} & 46.00 & \textbf{124} & 56.75 & 3,161 & 32.00 & 2,196\\
\hspace{0.15cm}w/o LoGiC in RL (\textit{RACE-SFT+GRPO}) & \textbf{75.54} & 2,573 & 51.40 & 2,670 & \textbf{67.18} & 8,574 & \textbf{35.75} & 3,507\\

\hspace{0.1cm}w/o LoGiC-RL (\textit{RACE-SFT}) & 43.05 & 1,966 & 45.40 & 1,232  & 63.08 & 3,330 & 28.25 & 1,856\\

\hspace{0.1cm}w/o LoGiC (\textit{SFT+GRPO}) & 74.55 & 2,073 & 52.00 & 2,702& 65.30 &3,545 & \textbf{35.75} & 2,163\\

\hspace{0.1cm}w/o Training (\textit{Base}) &  18.52 & 2,210 & 15.60 & 2,713 & 46.15 &3,134&29.25 & 1,281\\

\bottomrule
\end{tabular}
\end{table*}

\begin{figure*}[t]
    \centering 
    \includegraphics[width=0.98\linewidth]
    {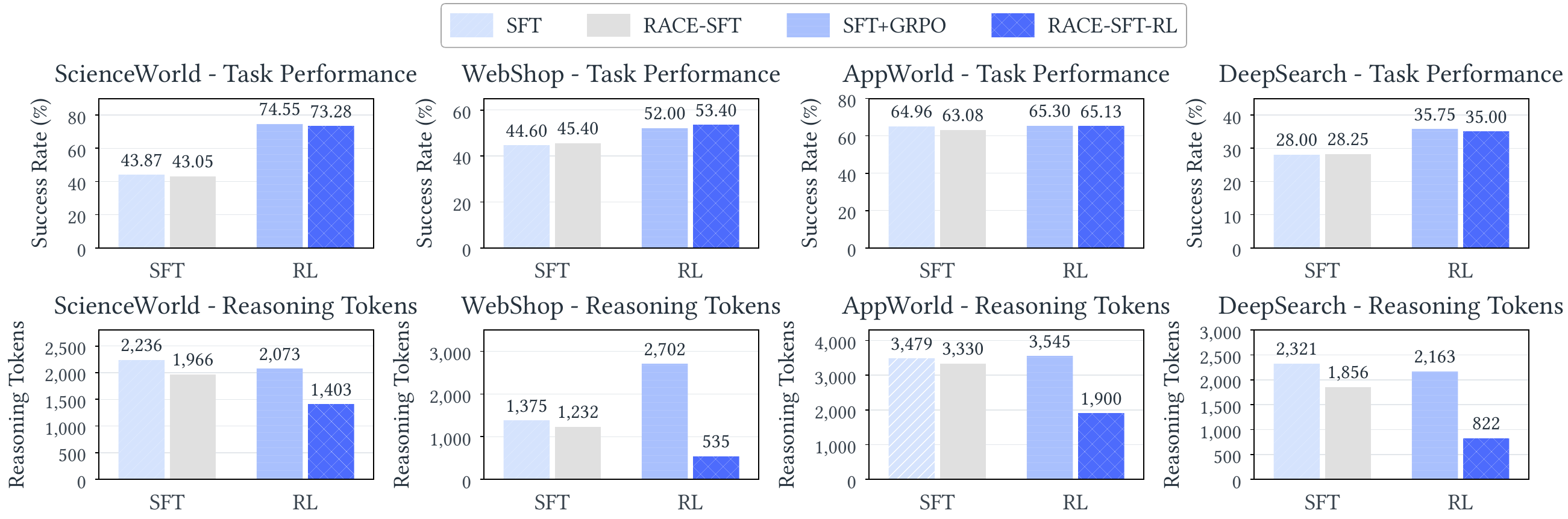}
    \caption{Comparison of task performance and reasoning efficiency with and without LoGiC.
    }
    \label{fig:analysis_compare}
\end{figure*}

\subsection{Quantitative Analysis}

\textbf{Ablation Studies.} 
We examine the effect of LoGiC-derived signals in SFT and RL. 
As shown in Figure~\ref{fig:analysis_compare}, RACE-SFT stays within 1.88 points of standard SFT, while reducing reasoning tokens by 4.3\% to 20.0\%. The efficiency gains are larger during RL. Compared with standard SFT+GRPO, RACE-SFT-RL stays within 1.40 points while reducing reasoning tokens by 32.3\%, 80.2\%, 46.4\%, and 62.0\% on ScienceWorld, WebShop, AppWorld, and DeepSearch, respectively. 
Table~\ref{tab:abalation} shows that removing LoGiC from RL sharply increases reasoning cost, from 1,403 to 2,573 tokens on ScienceWorld, 535 to 2,670 on WebShop, 1,900 to 8,574 on AppWorld, and 822 to 3,507 on DeepSearch. 
Removing reasoning masking also reduces success rate on ScienceWorld, WebShop, AppWorld, and DeepSearch, while its effect on reasoning cost varies across benchmarks.

\begin{figure*}[t]
    \centering 
    \includegraphics[width=0.98\linewidth]
    {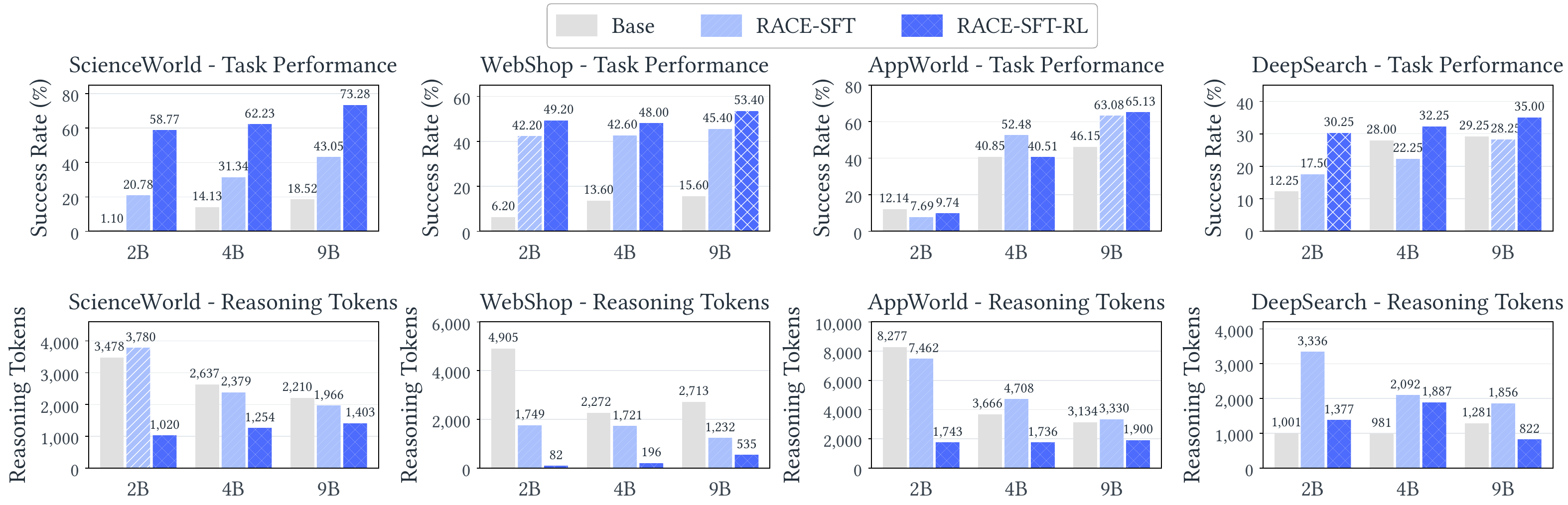}
    \caption{Scaling analysis with Qwen3.5-2B, 4B, and 9B across ScienceWorld, WebShop, AppWorld, and DeepSearch, showing task performance and reasoning efficiency at different model scales under no additional training, RACE-SFT, and RACE-SFT-RL.}
    \label{fig:scaling_analysis_success_reasoning}
\end{figure*}

\textbf{Scaling Analysis on the Qwen3.5 Series.}
We further analyze RACE across Qwen3.5-2B, 4B, and 9B on four agent benchmarks. As shown in Figure~\ref{fig:scaling_analysis_success_reasoning}, RACE generally achieves stronger task performance as model scale increases, although the gains vary across benchmarks.
On AppWorld, although performance improves substantially with model scale, the gains of RACE-SFT-RL over RACE-SFT vary across model sizes. This may be related to the limited training data available for AppWorld, which makes RL training less stable, especially at smaller model scales. Across model scales, RACE-SFT-RL generally improves task performance over RACE-SFT while also substantially reducing reasoning tokens, with especially large reductions on WebShop. Overall, these results further suggest that the effectiveness of RACE largely persists across different model scales.

\begin{wraptable}[13]{r}{0.5\linewidth}
    \vspace{-20pt}
    \centering
    \caption{Analysis of adaptive reasoning behavior. \textit{Skip} denotes the proportion of turns that directly execute actions without generating reasoning, while \textit{Tok./Think} denotes the average reasoning tokens over reasoning turns.}
    \label{tab:reasoning_behavior}
    \setlength{\tabcolsep}{2pt}
    \fontsize{8pt}{11pt}\selectfont
    \begin{tabular}{lcccc}
        \toprule
        \multirow{2}{*}{\textbf{Benchmark}}
        & \multicolumn{2}{c}{\textbf{RACE-SFT}}
        & \multicolumn{2}{c}{\textbf{RACE-SFT-RL}} \\
        \cmidrule(lr){2-3}
        \cmidrule(lr){4-5}
        & Skip (\%) & Tok./Think
        & Skip (\%) & Tok./Think \\
        \midrule
        ScienceWorld & 37.22 & 132 & 69.65 & 273 \\
        WebShop      & 7.15 & 144 & 55.72 & 101 \\
        AppWorld     & 7.74 & 273 & 52.15 & 302\\
        DeepSearch   & 1.79 & 259 & 5.77 & 117 \\
        \bottomrule
    \end{tabular}
\end{wraptable}

\textbf{Adaptive Reasoning Behavior Analysis.}
To understand how RACE reduces reasoning cost, we examine the skip ratio and average reasoning tokens over thinking turns. As shown in Table~\ref{tab:reasoning_behavior}, RL increases the skip ratio across all four benchmarks. Meanwhile, reasoning tokens per thinking turn increase on ScienceWorld and AppWorld but decrease on WebShop and DeepSearch. This suggests that RACE does not uniformly shorten reasoning. Instead, it skips reasoning at more turns, while retained reasoning adapts to task characteristics. In state-intensive tasks such as ScienceWorld and AppWorld, reasoning concentrates on more demanding interaction turns, whereas WebShop and DeepSearch favor more concise reasoning. 
Case studies are provided in Appendix~\ref{sec:case_study}.

\section{RELATED WORK}
\textbf{Reasoning-Efficient Methods.}
Prior studies have explored reducing unnecessary reasoning and improving reasoning efficiency in language models and agents~\citep{agentic_reasoning_survey}. One line reduces agent reasoning cost through more efficient planning, execution, or direct efficiency optimization~\citep{rewoo,tree_planner,depo,tool_light}. Another line studies adaptive reasoning more broadly, dynamically determining when or how much reasoning to perform~\citep{Gated_CoT,adaptthink,thinkless,adacot,arm,car,cprouter,ThinkSwitcher,ares,agent_omit,CogRouter}.
RACE extends this direction to turn-wise agent interaction, using a lightweight proxy to determine whether fresh reasoning is needed and applying the resulting signals to both SFT and agentic RL.

\textbf{Agentic Reinforcement Learning.}
Reinforcement learning (RL)~\citep{PPO, GRPO,DPO} is widely used to optimize agents from task-level feedback. Recent work has explored effective policy optimization~\citep{AEPO}, rollout exploration~\citep{ARPO,BPO}, and credit assignment~\citep{GiGPO,HGPO,graphgpo} for agent tasks. 
While these methods mainly focus on task performance, less attention has been paid to adaptive reasoning behavior during interaction. 
Our work complements this line by introducing adaptive reasoning into agentic reinforcement learning.

\section{Conclusion}
In this work, we present Reasoning Adaptation through Cross-Turn Estimation (RACE), a training approach for adaptive reasoning in LLM-based agents. Our analysis shows that earlier reasoning can support subsequent actions over varying ranges, while reference-action likelihood decrease provides a lightweight signal for estimating this support. Based on these findings, RACE introduces a Likelihood-Guided Progressive Reasoning Cover Detection (LoGiC) procedure to identify removable reasoning steps and use the resulting signals in both supervised fine-tuning and agentic reinforcement learning. Across four representative agent benchmarks, RACE substantially reduces reasoning cost while maintaining or improving task performance, demonstrating the effectiveness of likelihood-guided cross-turn estimation for learning efficient adaptive reasoning behavior.

\subsection*{AI use statement}
We used generative AI tools to assist with manuscript editing, including improving readability, refining technical descriptions, suggesting paper organization, and retrieving and discovering relevant work.
We have reviewed all AI-assisted work. We manually reviewed all AI-assisted content and verified that the revised text accurately reflects the authors' intended contributions and technical details.
We take responsibility for the final content of this work.


\subsection*{Reproducibility Statement}
We provide the complete workflow of the Likelihood-Guided Progressive Reasoning Cover Detection (LoGiC) in Appendix~\ref{sec:alg_LoGiC}. The datasets used in our experiments are described in Appendix~\ref{sec:datasets_details}, and the evaluation protocol and training configurations, including hyperparameters for SFT and RL, are provided in Appendices~\ref{sec:evaluation_protocol} and~\ref{sec:training_details}, respectively. The complete task-specific prompt templates for all four benchmarks are included in Appendix~\ref{sec:prompt_templates}.

\subsubsection*{Acknowledgments}
This work was supported by Alibaba Research Intern Program. We would like to thank the Qwen Team at Alibaba Token Hub (ATH), Alibaba Group, for providing the computational resources and foundation models (Qwen) used in this research.

\bibliography{iclr2027_conference}
\bibliographystyle{iclr2027_conference}

\newpage
\section*{Appendix}
\appendix
\startcontents[sections]
\printcontents[sections]{l}{1}{\setcounter{tocdepth}{3}}

\section{Tasks and Datasets}
\label{sec:datasets_details}
We evaluate our method on four representative agent tasks, covering scientific reasoning, web shopping, application use, and knowledge-intensive search. 
\subsection{Scientific Reasoning Task}
For scientific reasoning, we use \textbf{ScienceWorld}~\citep{sciworld}, a benchmark designed to evaluate agents on elementary-level science tasks in a text-based simulated environment. It comprises 30 tasks across 10 science topics: changes of state, temperature measurement, electrical circuits, friction, object classification, chemical mixtures, plants and pollinators, life spans, life stages, and Mendelian genetics. 
Each task requires the agent to accomplish a specified scientific goal through a sequence of textual observations and actions. We use the official training split for training and evaluate on the full official test split, which contains 1,819 task instances.

\subsection{Web Shopping Task}
For web shopping, we use \textbf{WebShop}~\citep{webshop}, which formulates shopping as a goal-directed decision-making task in the simulated e-commerce websites. WebShop contains 1,181,436 real-world products collected from Amazon and 12,087 crowd-sourced natural language instructions. Each instruction specifies requirements for a desired product, such as its attributes, options, and price, and the agent must search for relevant products, browse and compare candidates, select the required options, and complete the purchase. The instructions are officially divided into 10,587 training, 1,000 development, and 500 test instances.
We use the official training split for training and evaluate on the full test split of 500 instances.

\subsection{Application Use Task}
For application use, we adopt \textbf{AppWorld}~\citep{appworld}, where agents complete realistic day-to-day tasks by operating simulated applications through APIs. 
AppWorld provides 9 day-to-day applications with 457 APIs and includes 750 tasks instantiated from 250 task scenarios.
These tasks involve multiple applications and require agents to reason over intermediate results and adapt subsequent operations accordingly. Task completion is evaluated using state-based unit tests that verify the resulting application states.

We use both the official Train and Dev splits from the benchmark release for training. While the AppWorld paper reports 105 training tasks and 60 development tasks, the benchmark release used in our experiments contains 90 available Train tasks and 57 available Dev tasks. Therefore, our training set consists of 147 tasks in total.
For evaluation, we use the complete Test-N split with 168 tasks and the complete Test-C split with 417 tasks.

\subsection{Deep Search Task}
For deep search, we follow AgentGym-RL~\citep{agentgymrl} and consider queries from seven question answering datasets following the setup of Search-R1~\citep{search-r1}. These include \textbf{Natural Questions (NQ)}~\citep{NQ}, \textbf{TriviaQA}~\citep{TriviaQA}, and \textbf{PopQA}~\citep{PopQA} for general question answering, as well as \textbf{HotpotQA}~\citep{hotpotqa}, \textbf{2WikiMultiHopQA}~\citep{2WikiMultiHopQA}, \textbf{MuSiQue}~\citep{MuSiQue}, and \textbf{Bamboogle}~\citep{Bamboogle} for multi-hop question answering. Given a question, the agent can issue search queries to retrieve information and incorporate the retrieved evidence into its reasoning before producing the final answer.

We combine the training sets of NQ, HotpotQA, 2WikiMultiHopQA, and MuSiQue to construct the training data. Following AgentGym-RL, we evaluate on 400 examples randomly sampled from the development sets of seven datasets, covering diverse knowledge-intensive question answering tasks.

\section{Baseline Details}
\label{sec:baseline_details}
This section provides detailed descriptions of the baselines used in our experiments.

\subsection{ReAct-style Iterative Reasoning Agents}

\begin{itemize}[leftmargin=1em]
\item \textbf{DeepSeek-V4-Pro}~\citep{deepseek-v4} is a large-scale Mixture-of-Experts (MoE) language model released by the DeepSeek team. It consists of 1.6 trillion total parameters with 49 billion activated parameters and supports long-context processing. The model is designed for complex reasoning, coding, and agentic tasks, providing strong capabilities in handling challenging problems with extended contexts. We use DeepSeek-V4-Pro-0813 in our experiments.

\item \textbf{GLM-5.3}~\citep{glm-5} is an open-source large language model developed by Zhipu AI. It is designed for complex tasks including software engineering, reasoning, and agentic applications. The model supports long-context processing, function calling, and structured outputs, enabling it to handle multi-step workflows and interactive scenarios. 

\item \textbf{GPT-5.5}~\citep{openai2026gpt55} is OpenAI's frontier model for complex knowledge work and real-world problem solving. It provides capabilities in coding, information analysis, and tool use, with support for long-context processing and multi-step task execution across diverse scenarios. We evaluate GPT-5.5 with the reasoning effort set to medium.

\item \textbf{Claude-Opus-4.6}~\citep{anthropic2026claudeopus46} is a frontier model developed by Anthropic. It is designed for complex tasks including software engineering, research, analysis, and agentic workflows. The model supports extended thinking and long-context reasoning, enabling multi-step problem solving. We use Claude-Opus-4.6 in our experiments with the thinking level set to medium.

\item \textbf{Kimi-K3}~\citep{kimik3} is an open-weight multimodal model released by Moonshot AI. It adopts a Mixture-of-Experts architecture with 2.8 trillion total parameters, of which 104 billion are activated during inference. By incorporating Kimi Delta Attention (KDA) and Attention Residuals (AttnRes), the model supports multimodal understanding and contexts of up to one million tokens, making it suitable for complex reasoning and long-horizon agentic tasks.

\item \textbf{Qwen3.8 series}~\citep{qwen3.8} is an open-source large language model series developed by Alibaba's Qwen team. It is designed for a range of applications, including coding, professional tasks, research, and long-horizon agentic workflows. The series supports flexible thinking modes and agent execution, enabling the models to handle complex multi-step tasks. The series includes models of different scales to support diverse computational requirements and application scenarios. We use Qwen3.8-27B in our experiments.

\end{itemize}

\subsection{Reasoning-Efficient Agents}
\begin{itemize}[leftmargin=1em]

\item \textbf{Sparse ReAct}~\citep{react} is a prompting strategy based on the ReAct paradigm that reduces unnecessary reasoning generation during interaction. It allows the model to skip reasoning at some turns and directly execute actions when reasoning is not needed, while preserving the original ReAct interaction process between reasoning, actions, and observations.

\item \textbf{ADaPT}~\citep{adapt} is an approach for as-needed decomposition and planning in LLM-based agents. It adaptively performs task decomposition only when the executor cannot directly complete the current sub-task, avoiding unnecessary planning for simpler tasks. The approach employs a planner-executor framework and recursively decomposes complex tasks when needed. By adjusting the planning process according to task complexity, ADaPT reduces redundant decomposition and planning during agent execution.

\item \textbf{Think Tool}~\citep{anthropic_think_tool}
provides the agent with an auxiliary no-op tool for internal deliberation. The agent can autonomously decide whether to invoke this tool before taking subsequent actions. We implement this mechanism through prompting without
additional training.

\item \textbf{DART}~\citep{dart} adaptively selects between thinking and non-thinking modes based on the agreement of multiple no-think drafts. Since DART was originally designed for standalone query-level reasoning, we adapt it to our multi-step agent scenarios by sampling multiple no-thinking tool-call drafts at each interaction turn and using their agreement as a signal for deciding whether to invoke reasoning. This adaptation applies DART's decision process at the turn level, where the agent determines whether to generate reasoning before executing each action.

\item \textbf{AdaptThink}~\citep{adaptthink} learns to allocate reasoning by allowing models to choose between thinking and non-thinking modes. We extend this idea to interactive agents by making the reasoning-mode choice at every decision turn, where the agent either generates reasoning before an action or directly executes the action with an empty think block. Following the original formulation, we train the agent with GRPO using a modified reward $r(\tau)=R(\tau)+\delta\cdot\frac{1}{|\tau|}\sum_{t=1}^{|\tau|}\mathbb{I}(\mathrm{NoThink}_t)$, where the second term measures the proportion of non-thinking turns. The bonus term is applied only to successful trajectories, and we set $\delta=0.05$ while keeping the remaining GRPO settings unchanged.
\end{itemize}

\section{Implementation Details}
\label{sec:implementation_details}

\subsection{Evaluation Protocol}
\label{sec:evaluation_protocol}

We evaluate on ScienceWorld, WebShop, AppWorld, and DeepSearch. Across all benchmarks, we cap each episode at 30 interaction turns and allow up to 65,536 tokens for each model generation. We use a temperature of 1.0, top-$p$ of 1.0, and a maximum context length of 262,144 tokens, with reasoning enabled and native tool calling. ScienceWorld, WebShop, and DeepSearch allow at most one tool call per interaction turn, whereas AppWorld supports multiple API calls within a turn using precomputed and audited task-specific API lists. We report task success rate and the average total number of reasoning tokens per trajectory, measuring task performance and reasoning cost.

For ScienceWorld, we evaluate on 1,819 test instances spanning 30 task types, where an instance is considered successful only when the final environment score reaches 100. For WebShop, we use the official test split of 500 crowd-sourced instructions and consider a task successful when the environment reward reaches 1. For AppWorld, we evaluate on both Test-Normal and Test-Challenge and report the overall strict task success over all 585 tasks. DeepSearch is evaluated on 400 questions using Exact Match (EM).

\subsection{Training Details}
\label{sec:training_details}

\subsubsection{Supervised Fine-Tuning}

We construct the SFT data from successful trajectories annotated by DeepSeek-V4-Pro-0813. Specifically, we apply LoGiC with Qwen3.5-9B to these trajectories and set the likelihood-decrease threshold to $\epsilon=0.001$ to identify reasoning steps that can be removed. The resulting adaptive reasoning trajectories are then used to fine-tune Qwen3.5-9B. For ScienceWorld, WebShop, and DeepSearch, we set the learning rate to $1\times10^{-5}$ and the effective batch size to 32. Since AppWorld provides substantially fewer training trajectories, we use a smaller learning rate of $5\times10^{-6}$ and an effective batch size of 8. The maximum training sequence length is 65,536 tokens.

\subsubsection{Reinforcement Learning}

We initialize RL from the corresponding RACE-SFT checkpoint for each benchmark. We use a batch size of 16 and sample 8 trajectories per instance, yielding 128 trajectories per rollout batch. The learning rate is set to $1\times10^{-6}$, with one policy update epoch per rollout batch and a clipping ratio of 0.2. Rollouts are sampled with a temperature of 1.0 and top-$p$ of 1.0, with a maximum of 30 interaction turns.
LoGiC is applied only to successful on-policy trajectories during training, with the likelihood-decrease threshold set to $\epsilon=0.001$. We set the closure-loss weight to $\lambda=0.02$ for ScienceWorld, WebShop, and AppWorld, and $\lambda=0.01$ for DeepSearch.

\section{Prompt Template}
\label{sec:prompt_templates}
We use task-specific prompt templates for the four benchmarks, following their respective interaction protocols and tool interfaces. For AppWorld, we adopt the official prompt template. The full prompt templates are shown below.

\begin{tcolorbox}[
    enhanced jigsaw,
    breakable,
    pad at break*=1mm,
    colframe = gray!115,       
    colback = gray!5!white,             
    coltitle = white,                   
    coltext = black,                    
    fonttitle = \bfseries,              
    title = Prompt Template for ScienceWorld,  
    boxrule = 1pt,                      
    arc = 1.5mm,                          
    width = \linewidth,                 
    left = 7pt,                         
    right = 7pt,                        
    top = 5pt,                          
    bottom = 5pt                        
]
\fontsize{8.5pt}{10pt}\selectfont
You are an agent for science world. Every round I will give you an observation, and you must call exactly one of the provided functions to act toward finishing the given task. Only use the provided functions; the objects you choose must exist in the current room. If the environment returns ``No known action matches that input'', your previous action was invalid and you should try another option.
\end{tcolorbox}

\begin{tcolorbox}[
    enhanced jigsaw,
    breakable,
    pad at break*=1mm,
    colframe = gray!115,       
    colback = gray!5!white,             
    coltitle = white,                   
    coltext = black,                    
    fonttitle = \bfseries,              
    title = Prompt Template for WebShop,  
    boxrule = 1pt,                      
    arc = 1.5mm,                          
    width = \linewidth,                 
    left = 7pt,                         
    right = 7pt,                        
    top = 5pt,                          
    bottom = 5pt                        
]
\fontsize{8.5pt}{10pt}\selectfont
You are web shopping. I will give you instructions about what to do. Every round I will give you an observation and a list of available actions. Use the available tools to choose one valid action for the current page. Use search only if the search bar is available. Use click only with an exact value from the page clickables. If the page remains unchanged, it might indicate that your action is invalid.
\end{tcolorbox}

\begin{tcolorbox}[
    enhanced jigsaw,
    breakable,
    pad at break*=1mm,
    colframe = gray!115,       
    colback = gray!5!white,             
    coltitle = white,                   
    coltext = black,                    
    fonttitle = \bfseries,              
    title = Prompt Template for DeepSearch,  
    boxrule = 1pt,                      
    arc = 1.5mm,                          
    width = \linewidth,                 
    left = 7pt,                         
    right = 7pt,                        
    top = 5pt,                          
    bottom = 5pt                        
]
\fontsize{8.5pt}{10pt}\selectfont
Answer the question using your own knowledge and information returned by the search tool. \\
Use search only when necessary, call it at most once per turn, and avoid repeating equivalent searches. Once you have enough information, stop searching. \\
Only give me the final answer and do not output any other words. The final answer should be the shortest possible answer span, without explanations, Markdown, citations, or extra details.
\end{tcolorbox}

\clearpage
\begin{tcolorbox}[
    enhanced jigsaw,
    breakable,
    pad at break*=1mm,
    colframe = gray!115,       
    colback = gray!5!white,             
    coltitle = white,                   
    coltext = black,                    
    fonttitle = \bfseries,              
    title = Prompt Template for AppWorld,  
    boxrule = 1pt,                      
    arc = 1.5mm,                          
    width = \linewidth,                 
    left = 7pt,                         
    right = 7pt,                        
    top = 5pt,                          
    bottom = 5pt                        
]
\fontsize{8.5pt}{10pt}\selectfont
I am your supervisor, and you are an AI Assistant whose job is to complete my day-to-day tasks fully autonomously. \\
My name is: \texttt{\{full\_name\}}. My personal email is \texttt{\{email\}} and phone number is \texttt{\{phone\_number\}}.\\
You will be given a task instruction and a list of functions in the standard format. The functions correspond to APIs from various apps you have access to. The function name has two parts, the app name and API name separated by ``\_\_'', e.g., \texttt{spotify\_\_login} is the login API for the Spotify app.\\
You will complete the task completely autonomously through multi-turn interaction with the execution environment. In each turn, you will make one or more function calls, and the environment will return its outputs. This will continue either until you call the \texttt{complete\_task} API from the Supervisor app, or until a maximum of \texttt{\{max\_interactions\}} turns are reached. Here are brief app-wise descriptions. \texttt{\{app\_descriptions\}} \\
\textbf{Key Instructions:}

\textbf{A. General instructions:}

\begin{itemize}[leftmargin=1.5em]
    \item Act fully on your own. You must make all decisions yourself and never ask me or anyone else to confirm or clarify. Your role is to solve the task, not to bounce questions back, or provide me directions to follow.
    
    \item You have full access -- complete permission to operate across my connected accounts and services.
    
    \item Never invent or guess values. For example, if I ask you to play a song, do not assume the ID is 123. Instead, look it up properly through the right API.
    
    \item Never leave placeholders; don't output things like \texttt{your\_username}. Always fill in the real value by retrieving it via APIs (e.g., Supervisor app for credentials).
    
    \item When I omit details, choose any valid value. For example, if I ask you to buy something but don't specify which payment card to use, you may pick any one of my available cards.
    
    \item Avoid collateral damage. Only perform what I explicitly ask for. Example: if I ask you to buy something, do not delete emails, return the order, or perform unrelated account operations.
    
    \item You only have \texttt{\{max\_interactions\}} turns. Avoid unnecessary requests. You can batch unlimited function calls in a single turn -- always group them to save steps.
\end{itemize}

\textbf{B. App-specific instructions:}

\begin{itemize}[leftmargin=1.5em]
    \item All my personal information (biographical details, credentials, addresses, cards) is stored in the Supervisor app, accessible via its APIs.
    
    \item Any reference to my friends, family or any other person or relation refers to the people in my phone's contacts list.
    
    \item Always obtain current date or time, from the phone app's \texttt{get\_current\_date\_and\_time} API, never from your internal clock.
    
    \item All requests are concerning a single, default (no) time zone.
    
    \item For temporal requests, use proper time boundaries, e.g., when asked about periods like ``yesterday'', use complete ranges: \texttt{00:00:00} to \texttt{23:59:59}.
    
    \item References to ``file system'' mean the file system app, not the machine's OS. Do not use OS modules or functions.
    
    \item Paginated APIs: Always process all results, looping through the \texttt{page\_index}. Don't stop at the first page.
\end{itemize}

\textbf{C. Task-completion instructions:}

You must call the \texttt{supervisor\_\_complete\_task} API after completing the task.

\begin{itemize}[leftmargin=1.5em]
    \item The task is doable, but if you cannot find a way, you can call it with \texttt{status="fail"} to exit with failure.
\end{itemize}

When the answer is given:

\begin{itemize}[leftmargin=1.5em]
    \item Keep answers minimal. Return only the entity, number, or direct value requested -- not full sentences. E.g., for the song title of the current playing track, return just the title.
    
    \item Numbers must be numeric and not in words. E.g., for the number of songs in the queue, return ``10'', not ``ten''.
\end{itemize}

\end{tcolorbox}

\clearpage
\section{The Algorithm Workflow of the Likelihood-Guided Progressive Reasoning Cover Detection}
\label{sec:alg_LoGiC}
In this section, we present the detailed procedure of the Likelihood-Guided Progressive Reasoning Cover Detection (LoGiC) algorithm, as illustrated in Algorithm~\ref{alg:LoGiC}.

\begin{algorithm}[H]
\small
\caption{Likelihood-Guided Progressive Reasoning Cover Detection (LoGiC)}
\label{alg:LoGiC}
\textbf{Input} trajectory $\tau=(q,(r_t,a_t,o_t)_{t=1}^{T})$; 
reference actions $\{a_t^*\}_{t=1}^{T}$; 
likelihood function $\ell(\cdot)$; threshold $\epsilon$

\begin{algorithmic}[1]

\State Initialize action-likelihood vector 
$\bm{\ell}^{(0)}
=
[\ell(a_1^*|x_1,r_1),...,\ell(a_T^*|x_T,r_T)]$

\State Initialize compressed context $x^{(1)} \leftarrow q$, $x^{(2)} \leftarrow (x^{(1)},r_1,a_1^*,o_1)$
\State Initialize reasoning cover $\mathcal{C} \leftarrow \{r_1\}$
\State Initialize compressed trajectory 
$\tilde{\tau}\leftarrow(q,r_1,a_1^*,o_1)$

\For{$t=2,\dots,T$}

    \State Remove $r_t$ from the trajectory while keeping previous accepted removals, and construct the candidate trajectory:
    $
    (x^{(t)},a_t^*,o_t,r_{t+1},a_{t+1}^*,o_{t+1},...,r_T,a_T^*,o_T)
    $

    \State Compute candidate likelihood vector under the candidate trajectory:
    $\widetilde{\bm{\ell}}^{(t)}
    =
    [\widetilde{\ell}_1^{(t)},...,\widetilde{\ell}_T^{(t)}]$

    \State Compute likelihood decrease:
    $
    d_j^{(t)}
    =
    \ell_j^{(t-1)}
    -
    \widetilde{\ell}_j^{(t)},
    \quad
    j\in\{t,\dots,T\}
    $

    \If{$d_j^{(t)} \leq \epsilon,\ \forall j\in\{t,\dots,T\}$}

        \State Accept the removal of $r_t$

        \State Update likelihood vector:
        $\bm{\ell}^{(t)}
        \leftarrow
        \widetilde{\bm{\ell}}^{(t)}$

        \State Update compressed context:
        $x^{(t+1)}
        \leftarrow
        (x^{(t)},a_t^*,o_t)$

        \State Update compressed trajectory:
        $\tilde{\tau}
        \leftarrow
        (\tilde{\tau},a_t^*,o_t)$

    \Else

        \State Retain $r_t$

        \State Update likelihood vector:
        $\bm{\ell}^{(t)}
        \leftarrow
        \bm{\ell}^{(t-1)}$

        \State Update compressed context:
        $x^{(t+1)}
        \leftarrow
        (x^{(t)},r_t,a_t^*,o_t)$

        \State Update compressed trajectory:
        $\tilde{\tau}
        \leftarrow
        (\tilde{\tau},r_t,a_t^*,o_t)$

        \State Add $r_t$ into reasoning cover $\mathcal{C}$

    \EndIf

\EndFor

\State \textbf{Output} compressed trajectory $\tilde{\tau}$ and reasoning cover $\mathcal{C}$

\end{algorithmic}
\end{algorithm}

\section{Case Study}
\label{sec:case_study}
The examples in Tables~\ref{tab:case_sciworld}, \ref{tab:case_webshop}, \ref{tab:case_appworld}, and~\ref{tab:case_deepsearch} illustrate the adaptive reasoning behaviors learned through RACE across diverse agent tasks, including scientific reasoning, web shopping, application use, and deep search. Across these scenarios, the trained policy generates reasoning when needed while acting directly when the existing context already provides sufficient support for subsequent actions.

\subsection{Case Study for Scientific Reasoning}
\label{sec:case_sciworld}

Table~\ref{tab:case_sciworld} presents a ScienceWorld trajectory that requires combining red and blue paint to obtain violet paint. The policy model trained with RACE initially uses reasoning to infer the required composition and plan the manipulation sequence. Its first action, however, attempts to pick up the liquid directly and is rejected by the environment, revealing that the planned operation is infeasible.

The returned feedback triggers a short revision of the action strategy: instead of handling the liquid directly, the model switches to pouring it into the glass cup. Once this correction is established and the first pour succeeds, the remainder of the task becomes straightforward. The policy directly pours the blue paint, mixes the contents, and focuses on the resulting violet paint. This trajectory highlights how RACE training concentrates reasoning on planning and recovery from execution failure, while allowing the policy to carry out the remaining supported actions without additional reasoning.

\subsection{Case Study for Web Shopping}
\label{sec:case_webshop}

Table~\ref{tab:case_webshop} presents a WebShop trajectory in which a policy model trained with RACE adaptively allocates reasoning during product search and verification. The initial reasoning consolidates the user's constraints into a targeted query, leading to a candidate product that matches the requested item and price range. The policy then proceeds through the product page and description directly, since these navigation steps require little additional deliberation.

The need for reasoning emerges again only when the description fails to establish whether the product is individually wrapped. This missing attribute prompts the policy to inspect the feature page, where the remaining requirements are confirmed. With the product now fully verified, it returns to the product page and completes the purchase without further reasoning. The trajectory demonstrates how RACE training supports selective reasoning around uncertain product attributes while keeping routine browsing and selection steps concise.

\subsection{Case Study for Application Use}
\label{sec:case_appworld}

Table~\ref{tab:case_appworld} presents an AppWorld example that requires coordinated use of the file system, phone, and Gmail. The policy begins by decomposing the request into several subgoals: locating the birth certificate, finding the husband's contact information, and sending the document as an email attachment. With this workflow established, it carries out a sequence of routine operations, including account login and file retrieval, without generating reasoning at every turn.

Reasoning reappears when the task moves to a new subgoal that requires additional information or planning. After locating the target file, the policy recognizes that the husband's contact information is still needed and searches for the corresponding contact. After retrieving the email address, it reasons again to formulate the final email action and attach the correct birth certificate file. The remaining sending and task-completion steps are then executed directly without further reasoning. This trajectory highlights how RACE training leads the policy to concentrate reasoning around meaningful subgoal transitions, while avoiding unnecessary reasoning during routine operations whose next actions are already clear from the current context.

\subsection{Case Study for Deep Search}
\label{sec:case_deepsearch}

Table~\ref{tab:case_deepsearch} presents a DeepSearch example that requires identifying two film directors and comparing their birth dates. The policy first decomposes the question into the necessary information: the directors of \textit{Reaching for the Sun} and \textit{Trifling Women}, followed by their respective birth dates. Once this search structure is established, it retrieves the director of each film, skipping additional reasoning for the second lookup when the required action is already clear.

After both directors are identified, reasoning is invoked again to determine the remaining evidence needed for the comparison. The policy then retrieves their birth dates, directly issuing the second birth-date query after obtaining the first result. Finally, it combines the collected evidence and concludes that Rex Ingram, the director of \textit{Trifling Women}, was born earlier than William A. Wellman. The trajectory shows that RACE training encourages reasoning at points that require query decomposition or evidence synthesis, while allowing straightforward retrieval steps to proceed directly.

\clearpage
\begin{table*}[t]
    \centering
    \caption{A scientific reasoning example on ScienceWorld illustrating the adaptive reasoning behavior enabled by RACE across different interaction turns. [...] indicates omitted content. Special symbols highlight reasoning (\blue{blue box}), tool calls (\green{green box}), and tool results (\yellow{yellow box}).}
    \fontsize{9pt}{11pt}\selectfont
    \begin{tabular}{p{0.98\linewidth}}
    \midrule
        \rowcolor{gray!20}\textbf{Example from ScienceWorld Dataset} \\
    \midrule
        \textbf{Task Instruction} \\
        Your task is to use chemistry to create violet paint. When you are done, focus on the violet paint. \\
        \textbf{Initial Observation} \\
         This room is called the art studio. In it, you see: \\
        a substance called air \\
        a large cupboard. The large cupboard door is closed. \\
        a table. On the table is: a glass cup (containing nothing). \\
        a wood cup (containing red paint) \\
        a wood cup (containing blue paint) \\
        a wood cup (containing yellow paint)\\
  You also see:\\
        A door to the hallway (that is closed)\\
        \midrule
        \rowcolor{gray!20}
        \textbf{Model Output by RACE (9B)}\\
    \midrule
        \blue{<think>} We are in art studio. Need to create violet paint using chemistry. Violet is likely mixture of blue and red. We have red and blue paint. Need to combine to create violet. Maybe need proper containers. There is a glass cup and wood cups. Need to mix red and blue paints. The task says ``use chemistry to create violet paint. When you are done, focus on the violet paint.'' We need to use tool. Perhaps open? But first maybe move red paint and blue paint into glass cup, then mix. The red and blue are in wood cups. We can pickup? ``a wood cup (containing red paint)''. The object is ``red paint''? Maybe ``pick up red paint''? The action ``pickup'' moves object to inventory. Then use on glass cup? Actually ``mix'' mixes contents of container. We can move blue to glass cup, red to glass cup, then mix. But can we pick up paint? Likely ``pickup red paint'' then ``pickup blue paint'', then maybe ``move red paint to glass cup''? Inventory has both. Then move each to glass cup? Then mix. ...
        We can call ``pickup red paint'' to get red paint in inventory. Then ``pickup blue paint''. But ``red paint'' maybe ``wood cup''? Science word: objects are visible. ``a glass cup (containing nothing).'' ``a wood cup (containing red paint)''. We can call "pickup red paint"? If object not directly ``red paint''? It should be fine. Let's start by picking red paint. \blue{</think>}\\
        \green{<tool call>} pickup (obj="red paint") \green{</tool call>}  \\
        \yellow{<tool response>} You can't pick up a liquid directly. Try pouring it from one container to another, or dunking empty containers into containers filled with the liquid. \yellow{</tool response>} \\
        \blue{<think>} Need pour from wood cup into glass cup. We can ``pour red paint from wood cup to glass cup''? The tool ``pour'' likely ``liq'': ``red paint'', ``container'': ``glass cup''. But maybe need source? The L: ``pour a liquid into a container.'' So ``pour liquid into glass.'' They may require ``pour red paint into glass cup''? We can use ``pour'' with liq=``red paint'', container=``glass cup''. But if red paint in wood cup, pours from wood cup? Science word maybe pours from wherever. The object available is ``red paint''? Maybe it's ``a wood cup (containing red paint)''. The liquid is red paint. We can pour red paint into glass cup. Let's try. \blue{</think>}\\
        \green{<tool call>} pour(liq="red paint", container="glass cup") \green{</tool call>} \\
        \yellow{<tool response>} You pour the red paint into the glass cup. \yellow{</tool response>}\\
        \blue{<think>} \blue{</think>} \\
        \green{<tool call>} pour(liq="blue paint", container="glass cup") \green{</tool call>} \\
        \yellow{<tool response>} You pour the blue paint into the glass cup. \yellow{</tool response>} \\
        \blue{<think>} \blue{</think>} \\
        \green{<tool call>} mix(container="glass cup") \green{</tool call>} \\
        \yellow{<tool response>} Red and blue paint mix to produce violet paint. \yellow{</tool response>} \\
        \blue{<think>} \blue{</think>} \\
        \green{<tool call>} focus(obj="violet paint") \green{</tool call>} \\
        \yellow{<tool response>} You focus on the violet paint. \yellow{</tool response>} \\
    \bottomrule
    \end{tabular}
    \label{tab:case_sciworld} 
\end{table*}

\clearpage
\begin{table*}[t]
    \centering
    \caption{A web shopping example on WebShop illustrating the adaptive reasoning behavior enabled by RACE across different interaction turns, including product search, browsing, and selection. [...] indicates omitted content. Special symbols highlight reasoning (\blue{blue box}), tool calls (\green{green box}), tool results (\yellow{yellow box}), and clickable web elements (\gray{gray box}).}
    \fontsize{9pt}{11pt}\selectfont
    \begin{tabular}{p{0.98\linewidth}}
    \midrule
        \rowcolor{gray!20}\textbf{Example from WebShop Dataset} \\
    \midrule
        \textbf{Task Instruction} \\
        I like  traditional, old and individually wrapped albert's chocolate ice cubes 60 count tray chocolate, and price lower than 60.00 dollars \\
        \midrule
        \rowcolor{gray!20}
        \textbf{Model Output by RACE (9B)}\\
    \midrule
        \blue{<think>} The user wants to search for a product. Let me construct a search query based on the instruction. I'll search for ``Albert's Chocolate Ice Cubes 60 count tray traditional old individually wrapped''. \blue{</think>}\\
        \green{<tool call>} search (keywords="Albert's Chocolate Ice Cubes 60 count tray individually wrapped") \green{</tool call>}  \\
        \yellow{<tool response>} \\
        Page 1 (Total results: 50) \gray{Next >} \\
        \gray{B00T7JEX44} Albert's Chocolate Ice Cubes 60 Count Tray \$29.88 \\ 
        \gray{B00QOX3I6U} Albert's Chocolate Ice Cubes 50 Count \$19.91 \\
        \gray{B00A4BRAE4} Albert's Chocolate Ice Cubes 100 Count Tray \$29.62 ... \\
        \yellow{</tool response>} \\
        \blue{<think>} \blue{</think>} \\
        \green{<tool call>} click (item="B00T7JEX44") \green{</tool call>}  \\
        \yellow{<tool response>} \gray{Back to Search} \gray{< Prev} Albert's Chocolate Ice Cubes 60 Count Tray, Price: \$29.88, Rating: N.A. \gray{Description} \gray{Features} \gray{Reviews} \gray{Buy Now} \yellow{</tool response>}\\
        \blue{<think>} \blue{</think>} \\
        \green{<tool call>} click (item="Description") \green{</tool call>} \\
        \yellow{<tool response>} \gray{Back to Search} \gray{< Prev} Milk Chocolate Ice Cubes are the type of candy that chocolate lovers dream about. Because they are made from the highest quality ingredients, they are very perishable and have become quite hard to find.
        \yellow{</tool response>} \\
        \blue{<think>} \blue{</think>} \\
        \green{<tool call>} click (item="< prev") \green{</tool call>} \\
        \yellow{<tool response>} \gray{Back to Search}  \gray{< Prev} Albert's Chocolate Ice Cubes 60 Count Tray, Price: \$29.88, Rating: N.A. \gray{Description} \gray{Features} \gray{Reviews} \gray{Buy Now}
        \yellow{</tool response>}\\
        \blue{<think>} The description page doesn't mention individually wrapped. Let me check features. \blue{</think>} \\
        \green{<tool call>} click (item="Features") \green{</tool call>} \\
        \yellow{<tool response>} \gray{Back to Search} \gray{< Prev} Alberts Chocolate Ice Cubes - each container contains 60, 0.4 ounce, individually wrapped chocolate ice cubes. This is a traditional and old time favorite of many so order while you still can. These chocolate cubes are great for any candy dish or candy buffet and are sure to excite anyone looking for some old fashioned candy.
        \yellow{</tool response>}\\
        \blue{<think>} \blue{</think>} \\
        \green{<tool call>} click (item="< prev") \green{</tool call>} \\
        \yellow{<tool response>} \gray{Back to Search}  \gray{< Prev} Albert's Chocolate Ice Cubes 60 Count Tray, Price: \$29.88, Rating: N.A. \gray{Description} \gray{Features} \gray{Reviews} \gray{Buy Now}
        \yellow{</tool response>}\\
        \blue{<think>} \blue{</think>} \\
        \green{<tool call>} click (item="Buy Now") \green{</tool call>} \\
        \yellow{<tool response>} Thank you for shopping with us! \yellow{</tool response>} \\
    \bottomrule
    \end{tabular}
    \label{tab:case_webshop} 
\end{table*}

\begin{table*}[t]
    \centering
    \caption{An application use example on AppWorld illustrating the adaptive reasoning behavior enabled by RACE across different interaction turns. [...] indicates omitted content. Special symbols highlight reasoning (\blue{blue box}), tool calls (\green{green box}), and tool results (\yellow{yellow box}).}
    \fontsize{9pt}{11pt}\selectfont
    \begin{tabular}{p{0.98\linewidth}}
    \midrule
        \rowcolor{gray!20}\textbf{Example from AppWorld Dataset} \\
    \midrule
        \textbf{Task Instruction} \\
        Email the birth certificate found in my file system to my husband. \\
        \textbf{User Profile} \\
        My name is: Kristin White. My personal email is kri-powe@gmail.com and phone number is 6017026518.\\
 
        \midrule
        \rowcolor{gray!20}
        \textbf{Model Output by RACE (9B)}\\
    \midrule
        \blue{<think>} I need to email the birth certificate found in my file system to my husband. Let me start by understanding my accounts and finding the birth certificate file, then identify my husband's contact, and draft/send the email. First, I need credentials. Let me get supervisor profile and passwords, and maybe current phone/time. Let me call supervisor\_show\_account\_passwords and supervisor\_show\_profile.
        Also, maybe phone login requires phone number and password. I have personal email and phone number given: email kri-powe@gmail.com, phone 6017026518. Let me get credentials. Let me call supervisor APIs. \blue{</think>}\\
        \green{<tool call>} \\
        supervisor.show\_account\_passwords() \\
        supervisor.show\_profile() \\
        \green{</tool call>}  \\
        \yellow{<tool response>}  \\
        {[} \\
\quad \{"account\_name": "amazon", "password": "\$AMAZON\_PASSWORD"\}, \\
\quad \{"account\_name": "file\_system", "password": "\$FILE\_SYSTEM\_PASSWORD"\}, \\
\quad \{"account\_name": "gmail", "password": "\$GMAIL\_PASSWORD"\}, \\
\quad \{"account\_name": "phone", "password": "\$PHONE\_PASSWORD"\}, \\
\quad \{"account\_name": "simple\_note", "password": "\$SIMPLE\_NOTE\_PASSWORD"\}, \\
\quad \{"account\_name": "splitwise", "password": "\$SPLITWISE\_PASSWORD"\}, \\
\quad \{"account\_name": "spotify", "password": "\$SPOTIFY\_PASSWORD"\}, \\
\quad \{"account\_name": "todoist", "password": "\$TODOIST\_PASSWORD"\}, \\
\quad \{"account\_name": "venmo", "password": "\$VENMO\_PASSWORD"\} \\
] \\

        \yellow{</tool response>} \\ 
        \yellow{<tool response>} \\
        \{ \\
\quad "birthday": "1987-05-31", \\
\quad "email": "kri-powe@gmail.com", \\
\quad "first\_name": "Kristin", \\
\quad "last\_name": "White", \\
\quad "phone\_number": "6017026518", \\
\quad "sex": "female" \\
\}\\
 \yellow{</tool response>} \\ 
        \blue{<think>} \blue{</think>} \\
        \green{<tool call>} \\
        file\_system.login (username="kri-powe@gmail.com", password=FILE\_SYSTEM\_PASSWORD) \\
        phone.login (username="6017026518", password=PHONE\_PASSWORD) \\
        gmail.login (username="kri-powe@gmail.com", password=GMAIL\_PASSWORD) \\
        \green{</tool call>} \\
        \yellow{<tool response>} \\
        \{ "access\_token": "\$FILE\_SYSTEM\_TOKEN", "token\_type": "Bearer" \} \\
        \yellow{</tool response>} \\
        \yellow{<tool response>} \\
    \{ "access\_token": "\$PHONE\_TOKEN", "token\_type": "Bearer" \} \\
    \yellow{</tool response>} \\
    \yellow{<tool response>} \\
    \{ "access\_token": "\$GMAIL\_TOKEN", "token\_type": "Bearer" \} \\
     \yellow{</tool response>} \\
    \bottomrule
    \end{tabular}
    \label{tab:case_appworld} 
\end{table*}

\begin{table*}[t]
    \centering
    \fontsize{9pt}{11pt}\selectfont
    \begin{tabular}{p{0.98\linewidth}}
    \midrule
    \rowcolor{gray!20}
    \textbf{Example from AppWorld Dataset (continued)} \\
    \midrule


     \blue{<think>} \blue{</think>} \\
     \green{<tool call>} \\
      \mbox{file\_system.show\_directory(directory\_path="/", access\_token=FILE\_SYSTEM\_TOKEN, substring="birth")} \\
     \green{</tool call>} \\
     \yellow{<tool response>} \\
     "/home/kristin/documents/personal/identity/birth\_certificate.pdf" \\
     \yellow{</tool response>} \\
     \blue{<think>} Good, found birth certificate pdf. Now need to identify husband contact. Let me search contacts with relationship ``husband'' or query ``husband''. \blue{</think>} \\
     \green{<tool call>} \\
     phone.search\_contacts (relationship="husband", access\_token=PHONE\_TOKEN, page\_limit=20) \\
    \green{</tool call>} \\
    \yellow{<tool response>} \\
{[} \{ \\
\qquad "birthday": "1981-10-14", \\
\qquad "contact\_id": 540, \\
\qquad "created\_at": "2021-10-18T16:24:13", \\
\qquad "email": "alexwhite@gmail.com", \\
\qquad "first\_name": "Alex", \\
\qquad "home\_address": "0031 Jeremy Cliff Suite 552\textbackslash{}nSeattle\textbackslash{}nWashington\textbackslash{}nUnited States\textbackslash{}n72731", \\
\qquad "last\_name": "White", \\
\qquad "phone\_number": "3707865891", \\
\qquad "relationships": ["partner", "husband"], \\
\qquad "work\_address": "406 Lindsay Courts Suite 522\textbackslash{}nSeattle\textbackslash{}nWashington\textbackslash{}nUnited States\textbackslash{}n26281" \\
\quad  \} ] \\
    \yellow{</tool response>} \\
    \blue{<think>}
    Husband email is alexwhite@gmail.com. Now need to draft/send email with birth certificate attachment. Let me create draft with recipient alexwhite@gmail.com, subject maybe ``Birth Certificate'', body describing the document, and attach the file. The attachment file path is absolute: /home/kristin/documents/personal/identity/birth\_certificate.pdf. Need file\_system\_access\_token and gmail access\_token. Let me create draft with attachment\_file\_paths. The gmail create\_draft has optional scheduled\_send\_at etc, but we can just create draft and then send from draft. Let me create draft with attachment.
    \blue{</think>}\\
    \green{<tool call>} \\
    gmail.create\_draft(access\_token=GMAIL\_TOKEN, recipient\_email\_addresses=["alexwhite@gmail.com"], \\
    subject="Birth Certificate", body="Here is my birth certificate.", \\
    attachment\_file\_paths=["/home/kristin/documents/personal/identity/birth\_certificate.pdf"], file\_system\_access\_token=FILE\_SYSTEM\_TOKEN) \\
    \green{</tool call>} \\
    \yellow{<tool response>} \\
    \{ "draft\_id": 518, "message": "Draft created." \} \\
    \yellow{</tool response>} \\
    \blue{<think>} \blue{</think>} \\
    \green{<tool call>} \\
    gmail.send\_email\_from\_draft(draft\_id=518, access\_token=GMAIL\_TOKEN, \\
    file\_system\_access\_token=FILE\_SYSTEM\_TOKEN) \\
    \green{</tool call>} \\
    \yellow{<tool response>} \\
    \{ "message": "Sent email from your draft.", "sent\_email\_id": 20867, "sent\_email\_thread\_id": 47820 \}
    \yellow{</tool response>} \\
    \blue{<think>} \blue{</think>} \\
    \green{<tool call>} \\
    supervisor.complete\_task(status="success") \\
    \green{</tool call>} \\
    \yellow{<tool response>} \\
    \{ "message": "Marked the active task complete." \} \\
    \yellow{</tool response>} \\
    \bottomrule
    \end{tabular}
\end{table*}

\clearpage
\begin{table*}[t]
    \centering
    \caption{A deep search example on DeepSearch illustrating the adaptive reasoning behavior enabled by RACE across different interaction turns, including query decomposition, evidence collection, and answer synthesis. [...] indicates omitted content. Special symbols highlight reasoning (\blue{blue box}), tool calls (\green{green box}), and tool results (\yellow{yellow box}), and the final answer (\red{red box}).}
    \fontsize{9pt}{11pt}\selectfont
    \begin{tabular}{p{0.98\linewidth}}
    \midrule
        \rowcolor{gray!20}\textbf{Example from DeepSearch Dataset} \\
    \midrule
        \textbf{Task Instruction} \\
        Which film has the director who was born first, Reaching For The Sun or Trifling Women? \\
        
        \midrule
        \rowcolor{gray!20}
        \textbf{Model Output by RACE (9B)}\\
        \blue{<think>} We need answer question. Need identify directors of ``Reaching For The Sun'' and ``Trifling Women'', their birth dates, compare. Need search. Use search tool. \blue{</think>} \\
        \green{<tool call>}  search(query="Reaching For The Sun film director born") \green{</tool call>} \\
        \yellow{<tool response>} \\
        Doc 1(Title: "Reaching for the Sun") Reaching for the Sun Reaching for the Sun is a 1941 American comedy film directed by William A. Wellman and written by W.L. River. The film stars Joel McCrea, Ellen Drew, Eddie Bracken, Albert Dekker, Billy Gilbert, George Chandler and Bodil Ann Rosing... \\
        \yellow{</tool response>} \\
        \blue{<think>} \blue{</think>}\\
        \green{<tool call>} search(query='"Trifling Women" film director') \green{</tool call>}\\
        \yellow{<tool response>} \\
        Doc 1(Title: "Trifling Women") Trifling Women Trifling Women is a 1922 American silent romantic drama film directed by Rex Ingram. It is credited with boosting the careers of its leads, Barbara La Marr and Ramon Novarro. It has been described as Ingram's most personal film...\\
        \yellow{</tool response>} \\
        \blue{<think>} Need comparison of directors' birthdates. William A. Wellman? Rex Ingram? Need know ``Reaching For the Sun'' director William A. Wellman birth date. Let's search. Also need Rex Ingram birth date. Search. \blue{</think>} \\
        \green{<tool call>} search(query="William A. Wellman director birth date") \green{</tool call>}\\
        \yellow{<tool response>} \\
        Doc 1(Title: "William A. Wellman") William A. Wellman William Augustus Wellman (February 29, 1896 – December 9, 1975) was an American film director notable for his work in crime, adventure and action genre films, often focusing on aviation themes, a particular passion. He also directed several well-regarded satirical comedies...
        \yellow{</tool response>} \\
        \blue{<think>} \blue{</think>}\\
        \green{<tool call>} search(query="Rex Ingram director born birth date") \green{</tool call>}\\
        \yellow{<tool response>} \\
        Doc 1(Title: "Rex Ingram (director)") Rex Ingram (director) Rex Ingram (15 January 1892 – 21 July 1950) was an Irish film director, producer, writer and actor...\\
        \yellow{</tool response>} \\
        \blue{<think>} Reaching For the Sun director William A. Wellman born Feb 29, 1896. Trifling Women director Rex Ingram born Jan 15, 1892. Question "Which film has the director who was born first" -> Trifling Women (1892 < 1896). Final concise. \blue{</think>} \\
        \red{<answer}\\
        Trifling Women\\
        \red{</answer>} \\
        
    \bottomrule
    \end{tabular}
    \label{tab:case_deepsearch} 
\end{table*}

\end{document}

%% file: math_commands.tex
\usepackage{amsmath,amsfonts,bm}

\def\eqref#1{equation~\ref{#1}}

\def\1{\bm{1}}

\DeclareMathAlphabet{\mathsfit}{\encodingdefault}{\sfdefault}{m}{sl}
\SetMathAlphabet{\mathsfit}{bold}{\encodingdefault}{\sfdefault}{bx}{n}



%% file: iclr2027_conference.bib
@inproceedings{appworld,
  author       = {Harsh Trivedi and
                  Tushar Khot and
                  Mareike Hartmann and
                  Ruskin Manku and
                  Vinty Dong and
                  Edward Li and
                  Shashank Gupta and
                  Ashish Sabharwal and
                  Niranjan Balasubramanian},
  editor       = {Lun{-}Wei Ku and
                  Andre Martins and
                  Vivek Srikumar},
  title        = {AppWorld: {A} Controllable World of Apps and People for Benchmarking
                  Interactive Coding Agents},
  booktitle    = {Proceedings of the 62nd Annual Meeting of the Association for Computational
                  Linguistics (Volume 1: Long Papers), {ACL} 2024, Bangkok, Thailand,
                  August 11-16, 2024},
  pages        = {16022--16076},
  publisher    = {Association for Computational Linguistics},
  year         = {2024},
  url          = {https://doi.org/10.18653/v1/2024.acl-long.850},
  doi          = {10.18653/V1/2024.ACL-LONG.850},
  bibsource    = {dblp computer science bibliography, https://dblp.org}
}

@inproceedings{sciworld,
  author       = {Ruoyao Wang and
                  Peter A. Jansen and
                  Marc{-}Alexandre C{\^{o}}t{\'{e}} and
                  Prithviraj Ammanabrolu},
  editor       = {Yoav Goldberg and
                  Zornitsa Kozareva and
                  Yue Zhang},
  title        = {ScienceWorld: Is your Agent Smarter than a 5th Grader?},
  booktitle    = {Proceedings of the 2022 Conference on Empirical Methods in Natural
                  Language Processing, {EMNLP} 2022, Abu Dhabi, United Arab Emirates,
                  December 7-11, 2022},
  pages        = {11279--11298},
  publisher    = {Association for Computational Linguistics},
  year         = {2022},
  url          = {https://doi.org/10.18653/v1/2022.emnlp-main.775},
  doi          = {10.18653/V1/2022.EMNLP-MAIN.775},
  bibsource    = {dblp computer science bibliography, https://dblp.org}
}

@inproceedings{webshop,
  author       = {Shunyu Yao and
                  Howard Chen and
                  John Yang and
                  Karthik Narasimhan},
  editor       = {Sanmi Koyejo and
                  S. Mohamed and
                  A. Agarwal and
                  Danielle Belgrave and
                  K. Cho and
                  A. Oh},
  title        = {WebShop: Towards Scalable Real-World Web Interaction with Grounded
                  Language Agents},
  booktitle    = {Advances in Neural Information Processing Systems 35: Annual Conference
                  on Neural Information Processing Systems 2022, NeurIPS 2022, New Orleans,
                  LA, USA, November 28 - December 9, 2022},
  year         = {2022},
  url          = {http://papers.nips.cc/paper\_files/paper/2022/hash/82ad13ec01f9fe44c01cb91814fd7b8c-Abstract-Conference.html},
  bibsource    = {dblp computer science bibliography, https://dblp.org}
}

@article{search-r1,
  author       = {Bowen Jin and
                  Hansi Zeng and
                  Zhenrui Yue and
                  Dong Wang and
                  Hamed Zamani and
                  Jiawei Han},
  title        = {Search-R1: Training LLMs to Reason and Leverage Search Engines with
                  Reinforcement Learning},
  journal      = {CoRR},
  volume       = {abs/2503.09516},
  year         = {2025},
  url          = {https://doi.org/10.48550/arXiv.2503.09516},
  doi          = {10.48550/ARXIV.2503.09516},
  eprinttype   = {arXiv},
  eprint       = {2503.09516},
  bibsource    = {dblp computer science bibliography, https://dblp.org}
}

@inproceedings{hotpotqa,
  author       = {Zhilin Yang and
                  Peng Qi and
                  Saizheng Zhang and
                  Yoshua Bengio and
                  William W. Cohen and
                  Ruslan Salakhutdinov and
                  Christopher D. Manning},
  editor       = {Ellen Riloff and
                  David Chiang and
                  Julia Hockenmaier and
                  Jun'ichi Tsujii},
  title        = {HotpotQA: {A} Dataset for Diverse, Explainable Multi-hop Question
                  Answering},
  booktitle    = {Proceedings of the 2018 Conference on Empirical Methods in Natural
                  Language Processing, Brussels, Belgium, October 31 - November 4, 2018},
  pages        = {2369--2380},
  publisher    = {Association for Computational Linguistics},
  year         = {2018},
  url          = {https://doi.org/10.18653/v1/d18-1259},
  doi          = {10.18653/V1/D18-1259},
  bibsource    = {dblp computer science bibliography, https://dblp.org}
}

@article{agentgymrl,
  author       = {Zhiheng Xi and
                  Jixuan Huang and
                  Chenyang Liao and
                  Baodai Huang and
                  Honglin Guo and
                  Jiaqi Liu and
                  Rui Zheng and
                  Junjie Ye and
                  Jiazheng Zhang and
                  Wenxiang Chen and
                  Wei He and
                  Yiwen Ding and
                  Guanyu Li and
                  Zehui Chen and
                  Zhengyin Du and
                  Xuesong Yao and
                  Yufei Xu and
                  Jiecao Chen and
                  Tao Gui and
                  Zuxuan Wu and
                  Qi Zhang and
                  Xuanjing Huang and
                  Yu{-}Gang Jiang},
  title        = {AgentGym-RL: Training {LLM} Agents for Long-Horizon Decision Making
                  through Multi-Turn Reinforcement Learning},
  journal      = {CoRR},
  volume       = {abs/2509.08755},
  year         = {2025},
  url          = {https://doi.org/10.48550/arXiv.2509.08755},
  doi          = {10.48550/ARXIV.2509.08755},
  eprinttype   = {arXiv},
  eprint       = {2509.08755},
  bibsource    = {dblp computer science bibliography, https://dblp.org}
}

@article{NQ,
  author       = {Tom Kwiatkowski and
                  Jennimaria Palomaki and
                  Olivia Redfield and
                  Michael Collins and
                  Ankur P. Parikh and
                  Chris Alberti and
                  Danielle Epstein and
                  Illia Polosukhin and
                  Jacob Devlin and
                  Kenton Lee and
                  Kristina Toutanova and
                  Llion Jones and
                  Matthew Kelcey and
                  Ming{-}Wei Chang and
                  Andrew M. Dai and
                  Jakob Uszkoreit and
                  Quoc Le and
                  Slav Petrov},
  title        = {Natural Questions: a Benchmark for Question Answering Research},
  journal      = {Trans. Assoc. Comput. Linguistics},
  volume       = {7},
  pages        = {452--466},
  year         = {2019},
  url          = {https://doi.org/10.1162/tacl\_a\_00276},
  doi          = {10.1162/TACL\_A\_00276},
  bibsource    = {dblp computer science bibliography, https://dblp.org}
}

@inproceedings{TriviaQA,
  author       = {Mandar Joshi and
                  Eunsol Choi and
                  Daniel S. Weld and
                  Luke Zettlemoyer},
  editor       = {Regina Barzilay and
                  Min{-}Yen Kan},
  title        = {TriviaQA: {A} Large Scale Distantly Supervised Challenge Dataset for
                  Reading Comprehension},
  booktitle    = {Proceedings of the 55th Annual Meeting of the Association for Computational
                  Linguistics, {ACL} 2017, Vancouver, Canada, July 30 - August 4, Volume
                  1: Long Papers},
  pages        = {1601--1611},
  publisher    = {Association for Computational Linguistics},
  year         = {2017},
  url          = {https://doi.org/10.18653/v1/P17-1147},
  doi          = {10.18653/V1/P17-1147},
  bibsource    = {dblp computer science bibliography, https://dblp.org}
}

@inproceedings{PopQA,
  author       = {Alex Mallen and
                  Akari Asai and
                  Victor Zhong and
                  Rajarshi Das and
                  Daniel Khashabi and
                  Hannaneh Hajishirzi},
  editor       = {Anna Rogers and
                  Jordan L. Boyd{-}Graber and
                  Naoaki Okazaki},
  title        = {When Not to Trust Language Models: Investigating Effectiveness of
                  Parametric and Non-Parametric Memories},
  booktitle    = {Proceedings of the 61st Annual Meeting of the Association for Computational
                  Linguistics (Volume 1: Long Papers), {ACL} 2023, Toronto, Canada,
                  July 9-14, 2023},
  pages        = {9802--9822},
  publisher    = {Association for Computational Linguistics},
  year         = {2023},
  url          = {https://doi.org/10.18653/v1/2023.acl-long.546},
  doi          = {10.18653/V1/2023.ACL-LONG.546},
  bibsource    = {dblp computer science bibliography, https://dblp.org}
}

@inproceedings{2WikiMultiHopQA,
  author       = {Xanh Ho and
                  Anh{-}Khoa Duong Nguyen and
                  Saku Sugawara and
                  Akiko Aizawa},
  editor       = {Donia Scott and
                  N{\'{u}}ria Bel and
                  Chengqing Zong},
  title        = {Constructing {A} Multi-hop {QA} Dataset for Comprehensive Evaluation
                  of Reasoning Steps},
  booktitle    = {Proceedings of the 28th International Conference on Computational
                  Linguistics, {COLING} 2020, Barcelona, Spain (Online), December 8-13,
                  2020},
  pages        = {6609--6625},
  publisher    = {International Committee on Computational Linguistics},
  year         = {2020},
  url          = {https://doi.org/10.18653/v1/2020.coling-main.580},
  doi          = {10.18653/V1/2020.COLING-MAIN.580},
  bibsource    = {dblp computer science bibliography, https://dblp.org}
}

@article{MuSiQue,
  author       = {Harsh Trivedi and
                  Niranjan Balasubramanian and
                  Tushar Khot and
                  Ashish Sabharwal},
  title        = {MuSiQue: Multihop Questions via Single-hop Question
                  Composition},
  journal      = {Trans. Assoc. Comput. Linguistics},
  volume       = {10},
  pages        = {539--554},
  year         = {2022},
  url          = {https://doi.org/10.1162/tacl\_a\_00475},
  doi          = {10.1162/TACL\_A\_00475},
  bibsource    = {dblp computer science bibliography, https://dblp.org}
}

@inproceedings{Bamboogle,
  author       = {Ofir Press and
                  Muru Zhang and
                  Sewon Min and
                  Ludwig Schmidt and
                  Noah A. Smith and
                  Mike Lewis},
  editor       = {Houda Bouamor and
                  Juan Pino and
                  Kalika Bali},
  title        = {Measuring and Narrowing the Compositionality Gap in Language Models},
  booktitle    = {Findings of the Association for Computational Linguistics: {EMNLP}
                  2023, Singapore, December 6-10, 2023},
  series       = {Findings of {ACL}},
  volume       = {{EMNLP} 2023},
  pages        = {5687--5711},
  publisher    = {Association for Computational Linguistics},
  year         = {2023},
  url          = {https://doi.org/10.18653/v1/2023.findings-emnlp.378},
  doi          = {10.18653/V1/2023.FINDINGS-EMNLP.378},
  bibsource    = {dblp computer science bibliography, https://dblp.org}
}

@inproceedings{react,
  author       = {Shunyu Yao and
                  Jeffrey Zhao and
                  Dian Yu and
                  Nan Du and
                  Izhak Shafran and
                  Karthik R. Narasimhan and
                  Yuan Cao},
  title        = {ReAct: Synergizing Reasoning and Acting in Language Models},
  booktitle    = {The Eleventh International Conference on Learning Representations,
                  {ICLR} 2023, Kigali, Rwanda, May 1-5, 2023},
  publisher    = {OpenReview.net},
  year         = {2023},
  url          = {https://openreview.net/forum?id=WE\_vluYUL-X},
  bibsource    = {dblp computer science bibliography, https://dblp.org}
}

@article{rewoo,
  author       = {Binfeng Xu and
                  Zhiyuan Peng and
                  Bowen Lei and
                  Subhabrata Mukherjee and
                  Yuchen Liu and
                  Dongkuan Xu},
  title        = {ReWOO: Decoupling Reasoning from Observations for Efficient Augmented
                  Language Models},
  journal      = {CoRR},
  volume       = {abs/2305.18323},
  year         = {2023},
  url          = {https://doi.org/10.48550/arXiv.2305.18323},
  doi          = {10.48550/ARXIV.2305.18323},
  eprinttype   = {arXiv},
  eprint       = {2305.18323},
  bibsource    = {dblp computer science bibliography, https://dblp.org}
}

@inproceedings{adaptthink,
  author       = {Jiajie Zhang and
                  Nianyi Lin and
                  Lei Hou and
                  Ling Feng and
                  Juanzi Li},
  editor       = {Christos Christodoulopoulos and
                  Tanmoy Chakraborty and
                  Carolyn Rose and
                  Violet Peng},
  title        = {AdaptThink: Reasoning Models Can Learn When to Think},
  booktitle    = {Proceedings of the 2025 Conference on Empirical Methods in Natural
                  Language Processing, {EMNLP} 2025, Suzhou, China, November 4-9, 2025},
  pages        = {3716--3730},
  publisher    = {Association for Computational Linguistics},
  year         = {2025},
  url          = {https://doi.org/10.18653/v1/2025.emnlp-main.184},
  doi          = {10.18653/V1/2025.EMNLP-MAIN.184},
  bibsource    = {dblp computer science bibliography, https://dblp.org}
}

@article{dart,
  author       = {Jungseob Lee and
                  Seongtae Hong and
                  Seungjun Lee and
                  Jaehyung Seo and
                  Junyoung Son and
                  Sugyeong Eo and
                  Chanjun Park and
                  Hyeongju Park and
                  Hyeonseok Moon and
                  Heuiseok Lim},
  title        = {{DART:} Draft-Agreement Routing for Training-Free Adaptive Thinking
                  Budgets in Hybrid Reasoning Models},
  journal      = {CoRR},
  volume       = {abs/2606.23181},
  year         = {2026},
  url          = {https://doi.org/10.48550/arXiv.2606.23181},
  doi          = {10.48550/ARXIV.2606.23181},
  eprinttype   = {arXiv},
  eprint       = {2606.23181},
  bibsource    = {dblp computer science bibliography, https://dblp.org}
}

@article{car,
  author       = {Jinghui Lu and
                  Haiyang Yu and
                  Siliang Xu and
                  Shiwei Ran and
                  Guozhi Tang and
                  Siqi Wang and
                  Bin Shan and
                  Teng Fu and
                  Hao Feng and
                  Jingqun Tang and
                  Han Wang and
                  Can Huang},
  title        = {Prolonged Reasoning Is Not All You Need: Certainty-Based Adaptive
                  Routing for Efficient {LLM/MLLM} Reasoning},
  journal      = {CoRR},
  volume       = {abs/2505.15154},
  year         = {2025},
  url          = {https://doi.org/10.48550/arXiv.2505.15154},
  doi          = {10.48550/ARXIV.2505.15154},
  eprinttype   = {arXiv},
  eprint       = {2505.15154},
  bibsource    = {dblp computer science bibliography, https://dblp.org}
}

@inproceedings{cprouter,
  author       = {Jiayuan Su and
                  Fulin Lin and
                  Zhaopeng Feng and
                  Han Zheng and
                  Teng Wang and
                  Zhenyu Xiao and
                  Xinlong Zhao and
                  Zuozhu Liu and
                  Lu Cheng and
                  Hongwei Wang},
  editor       = {Sven Koenig and
                  Chad Jenkins and
                  Matthew E. Taylor},
  title        = {CP-Router: An Uncertainty-Aware Router Between {LLM} and {LRM}},
  booktitle    = {Fortieth {AAAI} Conference on Artificial Intelligence, Thirty-Eighth
                  Conference on Innovative Applications of Artificial Intelligence,
                  Sixteenth Symposium on Educational Advances in Artificial Intelligence,
                  {AAAI} 2026, Singapore, January 20-27, 2026},
  pages        = {33065--33073},
  publisher    = {{AAAI} Press},
  year         = {2026},
  url          = {https://doi.org/10.1609/aaai.v40i39.40589},
  doi          = {10.1609/AAAI.V40I39.40589},
  bibsource    = {dblp computer science bibliography, https://dblp.org}
}

@article{deepseek-v4,
  author       = {DeepSeek{-}AI},
  title        = {DeepSeek-V4: Towards Highly Efficient Million-Token Context Intelligence},
  journal      = {CoRR},
  volume       = {abs/2606.19348},
  year         = {2026},
  url          = {https://doi.org/10.48550/arXiv.2606.19348},
  doi          = {10.48550/ARXIV.2606.19348},
  eprinttype   = {arXiv},
  eprint       = {2606.19348},
  bibsource    = {dblp computer science bibliography, https://dblp.org}
}

@article{glm-5,
  author       = {GLM},
  title        = {{GLM-5:} from Vibe Coding to Agentic Engineering},
  journal      = {CoRR},
  volume       = {abs/2602.15763},
  year         = {2026},
  url          = {https://doi.org/10.48550/arXiv.2602.15763},
  doi          = {10.48550/ARXIV.2602.15763},
  eprinttype   = {arXiv},
  eprint       = {2602.15763},
  bibsource    = {dblp computer science bibliography, https://dblp.org}
}

@misc{openai2026gpt55,
  title        = {GPT-5.5 System Card},
  author       = {OpenAI},
  year         = {2026},
  howpublished = {\url{https://openai.com/index/gpt-5-5-system-card/}},
  note         = {Accessed: 2026-09-07}
}

@misc{anthropic2026claudeopus46,
  title        = {Introducing Claude Opus 4.6},
  author       = {Anthropic},
  year         = {2026},
  howpublished = {\url{https://www.anthropic.com/news/claude-opus-4-6}},
  note         = {Accessed: 2026-09-07}
}

@article{kimik3,
  author       = {{Kimi Team}},
  title        = {Kimi {K3:} Open Frontier Intelligence},
  journal      = {CoRR},
  volume       = {abs/2607.24653},
  year         = {2026},
  url          = {https://doi.org/10.48550/arXiv.2607.24653},
  doi          = {10.48550/ARXIV.2607.24653},
  eprinttype   = {arXiv},
  eprint       = {2607.24653},
  bibsource    = {dblp computer science bibliography, https://dblp.org}
}

@misc{qwen3.8,
    title  = {{Qwen3.8-Max}: A New Bar for Coding and Cowork},
    author = {{Qwen Team}},
    year   = {2026},
    month  = {August},
    url    = {https://qwen.ai/blog?id=qwen3.8},
    note   = {Accessed: 2026-09-25}
}

@misc{qwen3.5,
    title  = {{Qwen3.5}: Towards Native Multimodal Agents},
    author = {{Qwen Team}},
    year   = {2026},
    month  = {February},
    url    = {https://qwen.ai/blog?id=qwen3.5},
    note   = {Accessed: 2026-09-25}
}

@article{GRPO,
  author       = {Zhihong Shao and
                  Peiyi Wang and
                  Qihao Zhu and
                  Runxin Xu and
                  Junxiao Song and
                  Mingchuan Zhang and
                  Y. K. Li and
                  Y. Wu and
                  Daya Guo},
  title        = {DeepSeekMath: Pushing the Limits of Mathematical Reasoning in Open
                  Language Models},
  journal      = {CoRR},
  volume       = {abs/2402.03300},
  year         = {2024},
  url          = {https://doi.org/10.48550/arXiv.2402.03300},
  doi          = {10.48550/ARXIV.2402.03300},
  eprinttype   = {arXiv},
  eprint       = {2402.03300},
  bibsource    = {dblp computer science bibliography, https://dblp.org}
}

@inproceedings{depo,
  author       = {Sirui Chen and
                  Mengshi Zhao and
                  Lei Xu and
                  Yuying Zhao and
                  Beier Zhu and
                  Hanwang Zhang and
                  Shengjie Zhao and
                  Chaochao Lu},
  editor       = {Sven Koenig and
                  Chad Jenkins and
                  Matthew E. Taylor},
  title        = {{DEPO:} Dual-Efficiency Preference Optimization for {LLM} Agents},
  booktitle    = {Fortieth {AAAI} Conference on Artificial Intelligence, Thirty-Eighth
                  Conference on Innovative Applications of Artificial Intelligence,
                  Sixteenth Symposium on Educational Advances in Artificial Intelligence,
                  {AAAI} 2026, Singapore, January 20-27, 2026},
  pages        = {30279--30287},
  publisher    = {{AAAI} Press},
  year         = {2026},
  url          = {https://doi.org/10.1609/aaai.v40i36.40279},
  doi          = {10.1609/AAAI.V40I36.40279},
  bibsource    = {dblp computer science bibliography, https://dblp.org}
}

@inproceedings{tree_planner,
  author       = {Mengkang Hu and
                  Yao Mu and
                  Xinmiao Yu and
                  Mingyu Ding and
                  Shiguang Wu and
                  Wenqi Shao and
                  Qiguang Chen and
                  Bin Wang and
                  Yu Qiao and
                  Ping Luo},
  title        = {Tree-Planner: Efficient Close-loop Task Planning with Large Language
                  Models},
  booktitle    = {The Twelfth International Conference on Learning Representations,
                  {ICLR} 2024, Vienna, Austria, May 7-11, 2024},
  publisher    = {OpenReview.net},
  year         = {2024},
  url          = {https://openreview.net/forum?id=Glcsog6zOe},
  bibsource    = {dblp computer science bibliography, https://dblp.org}
}

@article{agentic_reasoning_survey,
  author       = {Tianxin Wei and
                  Ting{-}Wei Li and
                  Zhining Liu and
                  Xuying Ning and
                  Ze Yang and
                  Jiaru Zou and
                  Zhichen Zeng and
                  Ruizhong Qiu and
                  Xiao Lin and
                  Dongqi Fu and
                  Zihao Li and
                  Mengting Ai and
                  Duo Zhou and
                  Wenxuan Bao and
                  Yunzhe Li and
                  Gaotang Li and
                  Cheng Qian and
                  Yu Wang and
                  Xiangru Tang and
                  Yin Xiao and
                  Liri Fang and
                  Hui Liu and
                  Xianfeng Tang and
                  Yuji Zhang and
                  Chi Wang and
                  Jiaxuan You and
                  Heng Ji and
                  Hanghang Tong and
                  Jingrui He},
  title        = {Agentic Reasoning for Large Language Models},
  journal      = {CoRR},
  volume       = {abs/2601.12538},
  year         = {2026},
  url          = {https://doi.org/10.48550/arXiv.2601.12538},
  doi          = {10.48550/ARXIV.2601.12538},
  eprinttype   = {arXiv},
  eprint       = {2601.12538},
  bibsource    = {dblp computer science bibliography, https://dblp.org}
}

@article{tool_light,
  author       = {Yifei Chen and
                  Guanting Dong and
                  Zhicheng Dou},
  title        = {Toward Effective Tool-Integrated Reasoning via Self-Evolved Preference
                  Learning},
  journal      = {CoRR},
  volume       = {abs/2509.23285},
  year         = {2025},
  url          = {https://doi.org/10.48550/arXiv.2509.23285},
  doi          = {10.48550/ARXIV.2509.23285},
  eprinttype   = {arXiv},
  eprint       = {2509.23285},
  bibsource    = {dblp computer science bibliography, https://dblp.org}
}

@article{Gated_CoT,
  author       = {Samuel Lewis{-}Lim and
                  Xingwei Tan and
                  Zhixue Zhao and
                  Nikolaos Aletras},
  title        = {Can Confidence Estimates Decide When Chain-of-Thought Is Necessary
                  for LLMs?},
  journal      = {CoRR},
  volume       = {abs/2510.21007},
  year         = {2025},
  url          = {https://doi.org/10.48550/arXiv.2510.21007},
  doi          = {10.48550/ARXIV.2510.21007},
  eprinttype   = {arXiv},
  eprint       = {2510.21007},
  bibsource    = {dblp computer science bibliography, https://dblp.org}
}

@inproceedings{arm,
  author       = {Siye Wu and
                  Jian Xie and
                  Yikai Zhang and
                  Aili Chen and
                  Kai Zhang and
                  Yu Su and
                  Yanghua Xiao},
  editor       = {Danielle Belgrave and
                  Cheng Zhang and
                  Laura N. Montoya and
                  Hsuan{-}Tien Lin and
                  Razvan Pascanu and
                  Piotr Koniusz and
                  Marzyeh Ghassemi and
                  Nancy Chen and
                  Iv{\'{a}}n Vladimir Meza Ru{\'{\i}}z and
                  Arturo Loaiza{-}Bonilla},
  title        = {{ARM:} Adaptive Reasoning Model},
  booktitle    = {Advances in Neural Information Processing Systems 38: Annual Conference
                  on Neural Information Processing Systems 2025, NeurIPS 2025, San Diego,
                  CA, USA, December 2-7, 2025 / Mexico City, Mexico, November 30 - December
                  5, 2025},
  year         = {2025},
  url          = {http://papers.nips.cc/paper\_files/paper/2025/hash/48c218b01802a712ee8d7243d27fd9cf-Abstract-Conference.html},
  bibsource    = {dblp computer science bibliography, https://dblp.org}
}

@inproceedings{ThinkSwitcher,
  author       = {Guosheng Liang and
                  Longguang Zhong and
                  Ziyi Yang and
                  Xiaojun Quan},
  editor       = {Christos Christodoulopoulos and
                  Tanmoy Chakraborty and
                  Carolyn Rose and
                  Violet Peng},
  title        = {ThinkSwitcher: When to Think Hard, When to Think Fast},
  booktitle    = {Findings of the Association for Computational Linguistics: {EMNLP}
                  2025, Suzhou, China, November 4-9, 2025},
  pages        = {5185--5201},
  publisher    = {Association for Computational Linguistics},
  year         = {2025},
  url          = {https://doi.org/10.18653/v1/2025.findings-emnlp.278},
  doi          = {10.18653/V1/2025.FINDINGS-EMNLP.278},
  bibsource    = {dblp computer science bibliography, https://dblp.org}
}

@inproceedings{thinkless,
  author       = {Gongfan Fang and
                  Xinyin Ma and
                  Xinchao Wang},
  editor       = {Danielle Belgrave and
                  Cheng Zhang and
                  Laura N. Montoya and
                  Hsuan{-}Tien Lin and
                  Razvan Pascanu and
                  Piotr Koniusz and
                  Marzyeh Ghassemi and
                  Nancy Chen and
                  Iv{\'{a}}n Vladimir Meza Ru{\'{\i}}z and
                  Arturo Loaiza{-}Bonilla},
  title        = {Thinkless: {LLM} Learns When to Think},
  booktitle    = {Advances in Neural Information Processing Systems 38: Annual Conference
                  on Neural Information Processing Systems 2025, NeurIPS 2025, San Diego,
                  CA, USA, December 2-7, 2025 / Mexico City, Mexico, November 30 - December
                  5, 2025},
  year         = {2025},
  url          = {http://papers.nips.cc/paper\_files/paper/2025/hash/de2ad3ed44ee4e675b3be42aa0b615d0-Abstract-Conference.html},
  bibsource    = {dblp computer science bibliography, https://dblp.org}
}

@article{ares,
  author       = {Jingbo Yang and
                  Bairu Hou and
                  Wei Wei and
                  Yujia Bao and
                  Shiyu Chang},
  title        = {Ares: Adaptive Reasoning Effort Selection for Efficient {LLM} Agents},
  journal      = {CoRR},
  volume       = {abs/2603.07915},
  year         = {2026},
  url          = {https://doi.org/10.48550/arXiv.2603.07915},
  doi          = {10.48550/ARXIV.2603.07915},
  eprinttype   = {arXiv},
  eprint       = {2603.07915},
  bibsource    = {dblp computer science bibliography, https://dblp.org}
}

@article{adacot,
  author       = {Chenwei Lou and
                  Zewei Sun and
                  Xinnian Liang and
                  Meng Qu and
                  Wei Shen and
                  Wenqi Wang and
                  Yuntao Li and
                  Qingping Yang and
                  Shuangzhi Wu},
  title        = {AdaCoT: Pareto-Optimal Adaptive Chain-of-Thought Triggering via Reinforcement
                  Learning},
  journal      = {CoRR},
  volume       = {abs/2505.11896},
  year         = {2025},
  url          = {https://doi.org/10.48550/arXiv.2505.11896},
  doi          = {10.48550/ARXIV.2505.11896},
  eprinttype   = {arXiv},
  eprint       = {2505.11896},
  bibsource    = {dblp computer science bibliography, https://dblp.org}
}

@article{agent_omit,
  author       = {Yansong Ning and
                  Jun Fang and
                  Naiqiang Tan and
                  Hao Liu},
  title        = {Agent-Omit: Training Efficient {LLM} Agents for Adaptive Thought and
                  Observation Omission via Agentic Reinforcement Learning},
  journal      = {CoRR},
  volume       = {abs/2602.04284},
  year         = {2026},
  url          = {https://doi.org/10.48550/arXiv.2602.04284},
  doi          = {10.48550/ARXIV.2602.04284},
  eprinttype   = {arXiv},
  eprint       = {2602.04284},
  bibsource    = {dblp computer science bibliography, https://dblp.org}
}

@article{ARPO,
  author       = {Guanting Dong and
                  Hangyu Mao and
                  Kai Ma and
                  Licheng Bao and
                  Yifei Chen and
                  Zhongyuan Wang and
                  Zhongxia Chen and
                  Jiazhen Du and
                  Huiyang Wang and
                  Fuzheng Zhang and
                  Guorui Zhou and
                  Yutao Zhu and
                  Ji{-}Rong Wen and
                  Zhicheng Dou},
  title        = {Agentic Reinforced Policy Optimization},
  journal      = {CoRR},
  volume       = {abs/2507.19849},
  year         = {2025},
  url          = {https://doi.org/10.48550/arXiv.2507.19849},
  doi          = {10.48550/ARXIV.2507.19849},
  eprinttype   = {arXiv},
  eprint       = {2507.19849},
  bibsource    = {dblp computer science bibliography, https://dblp.org}
}

@inproceedings{GiGPO,
  author       = {Lang Feng and
                  Zhenghai Xue and
                  Tingcong Liu and
                  Bo An},
  editor       = {Danielle Belgrave and
                  Cheng Zhang and
                  Laura N. Montoya and
                  Hsuan{-}Tien Lin and
                  Razvan Pascanu and
                  Piotr Koniusz and
                  Marzyeh Ghassemi and
                  Nancy Chen and
                  Iv{\'{a}}n Vladimir Meza Ru{\'{\i}}z and
                  Arturo Loaiza{-}Bonilla},
  title        = {Group-in-Group Policy Optimization for {LLM} Agent Training},
  booktitle    = {Advances in Neural Information Processing Systems 38: Annual Conference
                  on Neural Information Processing Systems 2025, NeurIPS 2025, San Diego,
                  CA, USA, December 2-7, 2025 / Mexico City, Mexico, November 30 - December
                  5, 2025},
  year         = {2025},
  url          = {http://papers.nips.cc/paper\_files/paper/2025/hash/420c9f777c0b4f78d515e53cf74d58b2-Abstract-Conference.html},
  bibsource    = {dblp computer science bibliography, https://dblp.org}
}

@article{HGPO,
  author       = {Shuo He and
                  Lang Feng and
                  Qi Wei and
                  Xin Cheng and
                  Lei Feng and
                  Bo An},
  title        = {Hierarchy-of-Groups Policy Optimization for Long-Horizon Agentic Tasks},
  journal      = {CoRR},
  volume       = {abs/2602.22817},
  year         = {2026},
  url          = {https://doi.org/10.48550/arXiv.2602.22817},
  doi          = {10.48550/ARXIV.2602.22817},
  eprinttype   = {arXiv},
  eprint       = {2602.22817},
  bibsource    = {dblp computer science bibliography, https://dblp.org}
}

@article{graphgpo,
  author       = {Xin Cheng and
                  Shuo He and
                  Lang Feng and
                  Haiyang Xu and
                  Ming Yan and
                  Lei Feng and
                  Bo An},
  title        = {Beyond Trajectory-Level Attribution: Graph-Based Credit Assignment
                  for Agentic Reinforcement Learning},
  journal      = {CoRR},
  volume       = {abs/2605.26684},
  year         = {2026},
  url          = {https://doi.org/10.48550/arXiv.2605.26684},
  doi          = {10.48550/ARXIV.2605.26684},
  eprinttype   = {arXiv},
  eprint       = {2605.26684},
  bibsource    = {dblp computer science bibliography, https://dblp.org}
}

@inproceedings{deepagent,
  author       = {Xiaoxi Li and
                  Wenxiang Jiao and
                  Jiarui Jin and
                  Guanting Dong and
                  Jiajie Jin and
                  Yinuo Wang and
                  Hao Wang and
                  Yutao Zhu and
                  Ji{-}Rong Wen and
                  Yuan Lu and
                  Zhicheng Dou},
  editor       = {Hakim Hacid and
                  Yoelle Maarek and
                  Francesco Bonchi and
                  Ido Guy and
                  Emine Yilmaz},
  title        = {DeepAgent: {A} General Reasoning Agent with Scalable Toolsets},
  booktitle    = {Proceedings of the {ACM} Web Conference 2026, {WWW} 2026, Dubai, United
                  Arab Emirates, originally scheduled for April 13-17, 2026, rescheduled
                  for June 29 - July 3, 2026},
  pages        = {2219--2230},
  publisher    = {{ACM}},
  year         = {2026},
  url          = {https://doi.org/10.1145/3774904.3792460},
  doi          = {10.1145/3774904.3792460},
  bibsource    = {dblp computer science bibliography, https://dblp.org}
}

@article{PPO,
  author       = {John Schulman and
                  Filip Wolski and
                  Prafulla Dhariwal and
                  Alec Radford and
                  Oleg Klimov},
  title        = {Proximal Policy Optimization Algorithms},
  journal      = {CoRR},
  volume       = {abs/1707.06347},
  year         = {2017},
  url          = {http://arxiv.org/abs/1707.06347},
  eprinttype   = {arXiv},
  eprint       = {1707.06347},
  bibsource    = {dblp computer science bibliography, https://dblp.org}
}

@inproceedings{DPO,
  author       = {Rafael Rafailov and
                  Archit Sharma and
                  Eric Mitchell and
                  Christopher D. Manning and
                  Stefano Ermon and
                  Chelsea Finn},
  editor       = {Alice Oh and
                  Tristan Naumann and
                  Amir Globerson and
                  Kate Saenko and
                  Moritz Hardt and
                  Sergey Levine},
  title        = {Direct Preference Optimization: Your Language Model is Secretly a
                  Reward Model},
  booktitle    = {Advances in Neural Information Processing Systems 36: Annual Conference
                  on Neural Information Processing Systems 2023, NeurIPS 2023, New Orleans,
                  LA, USA, December 10 - 16, 2023},
  year         = {2023},
  url          = {http://papers.nips.cc/paper\_files/paper/2023/hash/a85b405ed65c6477a4fe8302b5e06ce7-Abstract-Conference.html},
  bibsource    = {dblp computer science bibliography, https://dblp.org}
}

@article{BPO,
  author       = {Bowei He and
                  Yankai Chen and
                  Xiaokun Zhang and
                  Xue Liu},
  title        = {Branching Policy Optimization: Sandbox-Native Language Agent Reinforcement
                  Learning},
  journal      = {CoRR},
  volume       = {abs/2607.14171},
  year         = {2026},
  url          = {https://doi.org/10.48550/arXiv.2607.14171},
  doi          = {10.48550/ARXIV.2607.14171},
  eprinttype   = {arXiv},
  eprint       = {2607.14171},
  bibsource    = {dblp computer science bibliography, https://dblp.org}
}

@article{AEPO,
  author       = {Guanting Dong and
                  Licheng Bao and
                  Zhongyuan Wang and
                  Kangzhi Zhao and
                  Xiaoxi Li and
                  Jiajie Jin and
                  Jinghan Yang and
                  Hangyu Mao and
                  Fuzheng Zhang and
                  Kun Gai and
                  Guorui Zhou and
                  Yutao Zhu and
                  Ji{-}Rong Wen and
                  Zhicheng Dou},
  title        = {Agentic Entropy-Balanced Policy Optimization},
  journal      = {CoRR},
  volume       = {abs/2510.14545},
  year         = {2025},
  url          = {https://doi.org/10.48550/arXiv.2510.14545},
  doi          = {10.48550/ARXIV.2510.14545},
  eprinttype   = {arXiv},
  eprint       = {2510.14545},
  bibsource    = {dblp computer science bibliography, https://dblp.org}
}

@article{chatshopbuddy,
  author       = {Yiruo Cheng and
                  Kelong Mao and
                  Tianhao Li and
                  Jiejun Tan and
                  Ji{-}Rong Wen and
                  Zhicheng Dou},
  title        = {ChatShopBuddy: Towards Reliable Conversational Shopping Agents via
                  Reinforcement Learning},
  journal      = {CoRR},
  volume       = {abs/2603.06065},
  year         = {2026},
  url          = {https://doi.org/10.48550/arXiv.2603.06065},
  doi          = {10.48550/ARXIV.2603.06065},
  eprinttype   = {arXiv},
  eprint       = {2603.06065},
  bibsource    = {dblp computer science bibliography, https://dblp.org}
}

@article{deepseek_r1,
  author       = {DeepSeek{-}AI},
  title        = {DeepSeek-R1: Incentivizing Reasoning Capability in LLMs via Reinforcement
                  Learning},
  journal      = {CoRR},
  volume       = {abs/2501.12948},
  year         = {2025},
  url          = {https://doi.org/10.48550/arXiv.2501.12948},
  doi          = {10.48550/ARXIV.2501.12948},
  eprinttype   = {arXiv},
  eprint       = {2501.12948},
  bibsource    = {dblp computer science bibliography, https://dblp.org}
}

@misc{qwq32b,
    title = {QwQ-32B: Embracing the Power of Reinforcement Learning},
    url = {https://qwen.ai/blog?id=qwq-32b},
     author = {{Qwen Team}},
    month = {March},
    year = {2025},
    note = {Accessed: 2026-09-25}
}

@misc{openai_o1,
  title  = {Learning to Reason with LLMs},
  author = {{OpenAI}},
  year   = {2024},
  month  = {September},
  url    = {https://openai.com/index/learning-to-reason-with-llms/},
  note   = {Accessed: 2026-09-25}
}

@inproceedings{webthinker,
  author       = {Xiaoxi Li and
                  Jiajie Jin and
                  Guanting Dong and
                  Hongjin Qian and
                  Yongkang Wu and
                  Ji{-}Rong Wen and
                  Yutao Zhu and
                  Zhicheng Dou},
  editor       = {Danielle Belgrave and
                  Cheng Zhang and
                  Laura N. Montoya and
                  Hsuan{-}Tien Lin and
                  Razvan Pascanu and
                  Piotr Koniusz and
                  Marzyeh Ghassemi and
                  Nancy Chen and
                  Iv{\'{a}}n Vladimir Meza Ru{\'{\i}}z and
                  Arturo Loaiza{-}Bonilla},
  title        = {WebThinker: Empowering Large Reasoning Models with Deep Research Capability},
  booktitle    = {Advances in Neural Information Processing Systems 38: Annual Conference
                  on Neural Information Processing Systems 2025, NeurIPS 2025, San Diego,
                  CA, USA, December 2-7, 2025 / Mexico City, Mexico, November 30 - December
                  5, 2025},
  year         = {2025},
  url          = {http://papers.nips.cc/paper\_files/paper/2025/hash/ae03bdef276132fae089692445725635-Abstract-Conference.html},
  bibsource    = {dblp computer science bibliography, https://dblp.org}
}

@article{ragen,
  author       = {Zihan Wang and
                  Kangrui Wang and
                  Qineng Wang and
                  Pingyue Zhang and
                  Linjie Li and
                  Zhengyuan Yang and
                  Xing Jin and
                  Kefan Yu and
                  Minh Nhat Nguyen and
                  Licheng Liu and
                  Eli Gottlieb and
                  Yiping Lu and
                  Kyunghyun Cho and
                  Jiajun Wu and
                  Li Fei{-}Fei and
                  Lijuan Wang and
                  Yejin Choi and
                  Manling Li},
  title        = {{RAGEN:} Understanding Self-Evolution in {LLM} Agents via Multi-Turn
                  Reinforcement Learning},
  journal      = {CoRR},
  volume       = {abs/2504.20073},
  year         = {2025},
  url          = {https://doi.org/10.48550/arXiv.2504.20073},
  doi          = {10.48550/ARXIV.2504.20073},
  eprinttype   = {arXiv},
  eprint       = {2504.20073},
  bibsource    = {dblp computer science bibliography, https://dblp.org}
}

@article{openresearcher,
  author       = {Zhuofeng Li and
                  Dongfu Jiang and
                  Xueguang Ma and
                  Haoxiang Zhang and
                  Ping Nie and
                  Yuyu Zhang and
                  Kai Zou and
                  Jianwen Xie and
                  Yu Zhang and
                  Wenhu Chen},
  title        = {OpenResearcher: {A} Fully Open Pipeline for Long-Horizon Deep Research
                  Trajectory Synthesis},
  journal      = {CoRR},
  volume       = {abs/2603.20278},
  year         = {2026},
  url          = {https://doi.org/10.48550/arXiv.2603.20278},
  doi          = {10.48550/ARXIV.2603.20278},
  eprinttype   = {arXiv},
  eprint       = {2603.20278},
  bibsource    = {dblp computer science bibliography, https://dblp.org}
}

@inproceedings{deepresearcher,
  author       = {Yuxiang Zheng and
                  Dayuan Fu and
                  Xiangkun Hu and
                  Xiaojie Cai and
                  Lyumanshan Ye and
                  Pengrui Lu and
                  Pengfei Liu},
  editor       = {Christos Christodoulopoulos and
                  Tanmoy Chakraborty and
                  Carolyn Rose and
                  Violet Peng},
  title        = {DeepResearcher: Scaling Deep Research via Reinforcement Learning in
                  Real-world Environments},
  booktitle    = {Proceedings of the 2025 Conference on Empirical Methods in Natural
                  Language Processing, {EMNLP} 2025, Suzhou, China, November 4-9, 2025},
  pages        = {414--431},
  publisher    = {Association for Computational Linguistics},
  year         = {2025},
  url          = {https://doi.org/10.18653/v1/2025.emnlp-main.22},
  doi          = {10.18653/V1/2025.EMNLP-MAIN.22},
  bibsource    = {dblp computer science bibliography, https://dblp.org}
}

@article{CogRouter,
  author       = {Ruihan Yang and
                  Fanghua Ye and
                  Xiang We and
                  Ruoqing Zhao and
                  Kang Luo and
                  Xinbo Xu and
                  Bo Zhao and
                  Ruotian Ma and
                  Shanyi Wang and
                  Zhaopeng Tu and
                  Xiaolong Li and
                  Deqing Yang and
                  Linus},
  title        = {Think Fast and Slow: Step-Level Cognitive Depth Adaptation for {LLM}
                  Agents},
  journal      = {CoRR},
  volume       = {abs/2602.12662},
  year         = {2026},
  url          = {https://doi.org/10.48550/arXiv.2602.12662},
  doi          = {10.48550/ARXIV.2602.12662},
  eprinttype   = {arXiv},
  eprint       = {2602.12662},
  bibsource    = {dblp computer science bibliography, https://dblp.org}
}

@article{search_reason,
  author       = {Qianben Chen and
                  Tianrui Qin and
                  King Zhu and
                  Qiexiang Wang and
                  Chengjun Yu and
                  Shu Xu and
                  Jiaqi Wu and
                  Jiayu Zhang and
                  Xinpeng Liu and
                  Xin Gui and
                  Jingyi Cao and
                  Piaohong Wang and
                  Dingfeng Shi and
                  He Zhu and
                  Tiannan Wang and
                  Yuqing Wang and
                  Maojia Song and
                  Tianyu Zheng and
                  Ge Zhang and
                  Jian Yang and
                  Jiaheng Liu and
                  Minghao Liu and
                  Yuchen Eleanor Jiang and
                  Wangchunshu Zhou},
  title        = {Search More, Think Less: Rethinking Long-Horizon Agentic Search for
                  Efficiency and Generalization},
  journal      = {CoRR},
  volume       = {abs/2602.22675},
  year         = {2026},
  url          = {https://doi.org/10.48550/arXiv.2602.22675},
  doi          = {10.48550/ARXIV.2602.22675},
  eprinttype   = {arXiv},
  eprint       = {2602.22675},
  bibsource    = {dblp computer science bibliography, https://dblp.org}
}

@inproceedings{steer,
  author       = {Sangmook Lee and
                  Dohyung Kim and
                  Hyukhun Koh and
                  Nakyeong Yang and
                  Kyomin Jung},
  editor       = {Sven Koenig and
                  Chad Jenkins and
                  Matthew E. Taylor},
  title        = {Confidence-Guided Stepwise Model Routing for Cost-Efficient Reasoning},
  booktitle    = {Fortieth {AAAI} Conference on Artificial Intelligence, Thirty-Eighth
                  Conference on Innovative Applications of Artificial Intelligence,
                  Sixteenth Symposium on Educational Advances in Artificial Intelligence,
                  {AAAI} 2026, Singapore, January 20-27, 2026},
  pages        = {31483--31491},
  publisher    = {{AAAI} Press},
  year         = {2026},
  url          = {https://doi.org/10.1609/aaai.v40i37.40413},
  doi          = {10.1609/AAAI.V40I37.40413},
  bibsource    = {dblp computer science bibliography, https://dblp.org}
}

@inproceedings{adapt,
  author       = {Archiki Prasad and
                  Alexander Koller and
                  Mareike Hartmann and
                  Peter Clark and
                  Ashish Sabharwal and
                  Mohit Bansal and
                  Tushar Khot},
  editor       = {Kevin Duh and
                  Helena G{\'{o}}mez{-}Adorno and
                  Steven Bethard},
  title        = {ADaPT: As-Needed Decomposition and Planning with Language Models},
  booktitle    = {Findings of the Association for Computational Linguistics: {NAACL}
                  2024, Mexico City, Mexico, June 16-21, 2024},
  series       = {Findings of {ACL}},
  volume       = {{NAACL} 2024},
  pages        = {4226--4252},
  publisher    = {Association for Computational Linguistics},
  year         = {2024},
  url          = {https://doi.org/10.18653/v1/2024.findings-naacl.264},
  doi          = {10.18653/V1/2024.FINDINGS-NAACL.264},
  bibsource    = {dblp computer science bibliography, https://dblp.org}
}

@misc{anthropic_think_tool,
  author = {Anthropic},
  title = {The ``think'' Tool: Enabling Claude to Stop and Think in Complex Tool Use Situations},
  year = {2025},
  howpublished = {\url{https://www.anthropic.com/engineering/claude-think-tool}},
  note         = {Accessed: 2026-09-25}
}

@article{htr,
  author       = {Jiajie Jin and
                  Yuyang Hu and
                  Kai Qiu and
                  Qi Dai and
                  Chong Luo and
                  Guanting Dong and
                  Xiaoxi Li and
                  Tong Zhao and
                  Xiaolong Ma and
                  Gongrui Zhang and
                  Zhirong Wu and
                  Bei Liu and
                  Zhengyuan Yang and
                  Linjie Li and
                  Lijuan Wang and
                  Hongjin Qian and
                  Yutao Zhu and
                  Zhicheng Dou},
  title        = {Toward Generalist Autonomous Research via Hypothesis-Tree Refinement},
  journal      = {CoRR},
  volume       = {abs/2606.11926},
  year         = {2026},
  url          = {https://doi.org/10.48550/arXiv.2606.11926},
  doi          = {10.48550/ARXIV.2606.11926},
  eprinttype   = {arXiv},
  eprint       = {2606.11926},
  bibsource    = {dblp computer science bibliography, https://dblp.org}
}

@inproceedings{hiersearch,
  author       = {Jiejun Tan and
                  Zhicheng Dou and
                  Yan Yu and
                  Jiehan Cheng and
                  Lifeng Liu and
                  Jian Xie and
                  Jirong Wen},
  editor       = {Sven Koenig and
                  Chad Jenkins and
                  Matthew E. Taylor},
  title        = {HierSearch: {A} Hierarchical Enterprise Deep Search Framework Integrating
                  Local and Web Searches},
  booktitle    = {Fortieth {AAAI} Conference on Artificial Intelligence, Thirty-Eighth
                  Conference on Innovative Applications of Artificial Intelligence,
                  Sixteenth Symposium on Educational Advances in Artificial Intelligence,
                  {AAAI} 2026, Singapore, January 20-27, 2026},
  pages        = {19380--19388},
  publisher    = {{AAAI} Press},
  year         = {2026},
  url          = {https://doi.org/10.1609/aaai.v40i23.39015},
  doi          = {10.1609/AAAI.V40I23.39015},
  bibsource    = {dblp computer science bibliography, https://dblp.org}
}
